\documentclass[10pt,twocolumn,letterpaper]{article}

\usepackage[pagenumbers]{wacv} %

\usepackage{graphicx}
\usepackage{amsmath}
\usepackage{amssymb}
\usepackage{booktabs}
\usepackage{multirow}
\usepackage{floatrow}

\usepackage[pagebackref,breaklinks,colorlinks]{hyperref}

\usepackage[capitalize]{cleveref}
\crefname{section}{Sec.}{Secs.}
\Crefname{section}{Section}{Sections}
\Crefname{table}{Table}{Tables}
\crefname{table}{Tab.}{Tabs.}

\def\wacvPaperID{2203} %
\def\confName{WACV}
\def\confYear{2024}

\begin{document}

\title{Beard Segmentation and Recognition Bias}

\author{Kagan Ozturk, Grace Bezold, Aman Bhatta, Haiyu Wu, Kevin Bowyer\\
Computer Vision Research Lab\\
University of Notre Dame\\
{\tt\small \{kztrk, gbezold, abhatta, hwu6, kwb\}@nd.edu}\\
}

\maketitle

\begin{abstract}

A person’s facial hairstyle, such as presence and size of beard, can significantly impact face recognition accuracy. There are publicly-available deep networks that achieve reasonable accuracy at binary attribute classification, such as beard / no beard, but few if any that segment the facial hair region. To investigate the effect of facial hair in a rigorous manner, we first created a set of fine-grained facial hair annotations to train a segmentation model and evaluate its accuracy across African-American and Caucasian face images. We then use our facial hair segmentations to categorize image pairs according to the degree of difference or similarity in the facial hairstyle. We find that the False Match Rate (FMR) for image pairs with different categories of facial hairstyle varies by a factor of over 10 for African-American males and over 25 for Caucasian males. To reduce the bias across image pairs with different facial hairstyles, we propose a scheme for adaptive thresholding based on facial hairstyle similarity. Evaluation on a subject-disjoint set of images shows that adaptive similarity thresholding based on facial hairstyles of the image pair reduces the ratio between the highest and lowest FMR across facial hairstyle categories for African-American from 10.7 to 1.8 and for Caucasians from 25.9 to 1.3. Facial hair annotations and facial hair segmentation model will be publicly available.

\end{abstract}

\section{Introduction}
\label{sec:intro}

Recent developments in deep neural networks (DNNs) have enabled learning complex feature hierarchies from raw data. With the availability of large datasets and computational power, end-to-end representation learning has become the most prominent method in computer vision as well as many other domains. Face recognition has been one of the most popular computer vision problems for researchers over the past decades, and 
face recognition has found extensive use in various real-world applications. %

Accuracy of face verification improved with the advent of DNNs \cite{adaface, arcface, deep_face, cosface, schroff2015facenet}. However,  studies have demonstrated that the accuracy of these models varies between demographic groups \cite{grother2019face, terhorst2021comprehensive, thong2021feature, wu2022face, serna2019algorithmic, 9086771, 9209125, abdurrahim2018review, klare2012face}. Although the initial approach to identify the root cause of these disparities involves examination of training data, even when the data is balanced among demographic groups, the models may still yield different impostor (images of two different individuals) and genuine (different images of the same individual) distributions \cite{Albiero2020training, georgopoulos2021mitigating, Xu_2021_CVPR}. 

Beyond demographics, several face attributes, including facial hair, are investigated for possible causes of bias in face recognition (see Section \ref{sec:related_work}). Facial hair can readily be used to change appearance in a significant way; e.g., 
a person can shave their long beard to have a clean-shaven face. At the same time, compared to other occlusions such as masks, sunglasses, hat, etc., one can be requested to take off their face mask to take a border-crossing photo, but it is not feasible to ask someone to shave at the border-crossing kiosk.

\begin{figure*}[htb]
    \captionsetup[subfigure]{aboveskip=2pt,belowskip=5pt}
    \begin{subfigure}[t]{\textwidth}
        \begin{subfigure}[t]{0.49\textwidth}
            \centering
            \begin{subfigure}[t]{0.32\textwidth}
                \centering
                \includegraphics[width=1\linewidth]{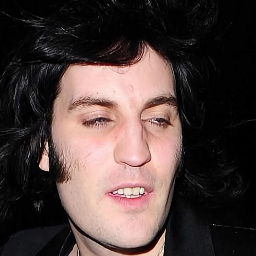}
            \end{subfigure}
            \begin{subfigure}[t]{0.32\textwidth}
                \centering
                \includegraphics[width=1\linewidth]{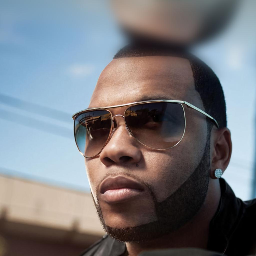}
            \end{subfigure}
            \begin{subfigure}[t]{0.32\textwidth}
                \centering
                \includegraphics[width=1\linewidth]{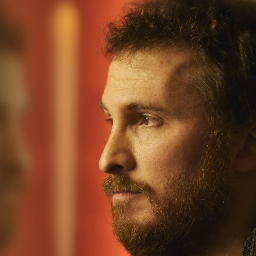}
            \end{subfigure}
        \end{subfigure}
        \begin{subfigure}[t]{0.49\textwidth}
            \centering
            \begin{subfigure}[t]{0.32\textwidth}
                \centering
                \includegraphics[width=1\linewidth]{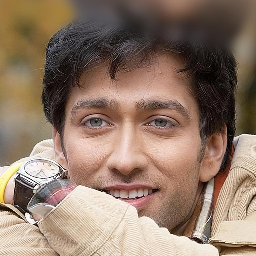}
            \end{subfigure}
            \begin{subfigure}[t]{0.32\textwidth}
                \centering
                \includegraphics[width=1\linewidth]{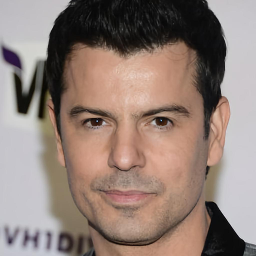}
            \end{subfigure}
            \begin{subfigure}[t]{0.32\textwidth}
                \centering
                \includegraphics[width=1\linewidth]{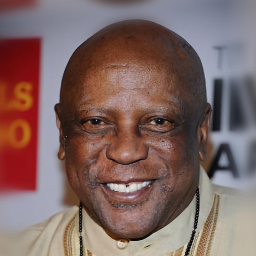}
            \end{subfigure}
        \end{subfigure}
        \caption{CelebA-HQ images}
    \end{subfigure}

    \begin{subfigure}[t]{\textwidth}
        \begin{subfigure}[t]{0.49\textwidth}
            \centering
            \begin{subfigure}[t]{0.32\textwidth}
                \centering
                \includegraphics[width=1\linewidth]{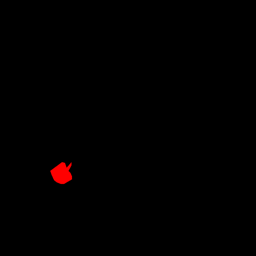}
            \end{subfigure}
            \begin{subfigure}[t]{0.32\textwidth}
                \centering
                \includegraphics[width=1\linewidth]{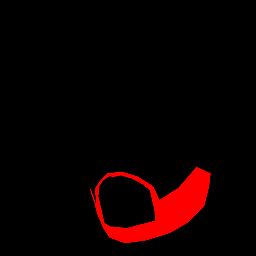}
            \end{subfigure}
            \begin{subfigure}[t]{0.32\textwidth}
                \centering
                \includegraphics[width=1\linewidth]{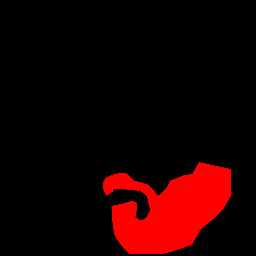}
            \end{subfigure}
            \caption*{Facial Hair (Red)}
        \end{subfigure}
        \begin{subfigure}[t]{0.49\textwidth}
            \centering
            \begin{subfigure}[t]{0.32\textwidth}
                \centering
                \includegraphics[width=1\linewidth]{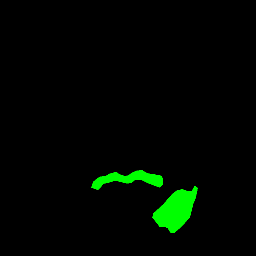}
            \end{subfigure}
            \begin{subfigure}[t]{0.32\textwidth}
                \centering
                \includegraphics[width=1\linewidth]{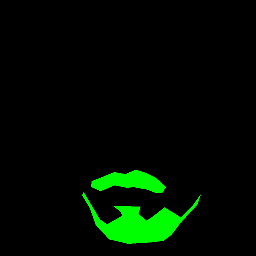}
            \end{subfigure}
            \begin{subfigure}[t]{0.32\textwidth}
                \centering
                \includegraphics[width=1\linewidth]{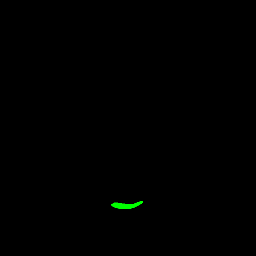}
            \end{subfigure}
            \caption*{Five o'clock shadows (Green)}
        \end{subfigure}
        \vspace{-8pt}
        \caption{Hand-annotations}   
    \end{subfigure}

    \begin{subfigure}[t]{\textwidth}
        \begin{subfigure}[t]{0.49\textwidth}
            \centering
            \begin{subfigure}[t]{0.32\textwidth}
                \includegraphics[width=1\linewidth]{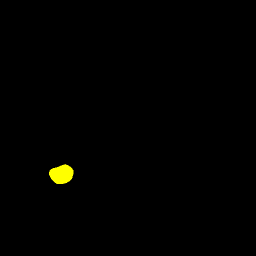}
                \caption*{IoU: 78.08}
            \end{subfigure}
            \begin{subfigure}[t]{0.32\textwidth}
                \centering
                \includegraphics[width=1\linewidth]{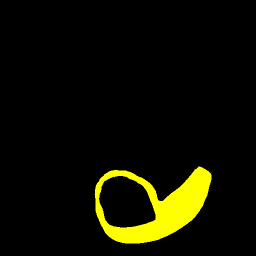}
                \caption*{IoU: 88.04}
            \end{subfigure}
            \begin{subfigure}[t]{0.32\textwidth}
                \centering
                \includegraphics[width=1\linewidth]{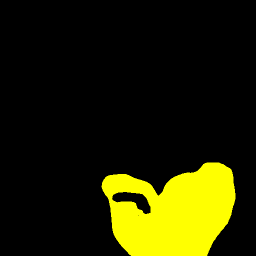}
                \caption*{IoU: 81.09}
            \end{subfigure}
        \end{subfigure}
        \begin{subfigure}[t]{0.49\textwidth}
            \centering
            \begin{subfigure}[t]{0.32\textwidth}
                \centering
                \includegraphics[width=1\linewidth]{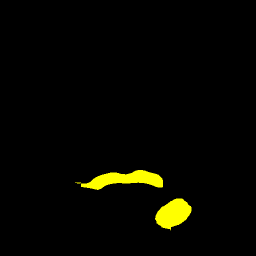}
                \caption*{IoU: 67.74}
            \end{subfigure}
            \begin{subfigure}[t]{0.32\textwidth}
                \centering
                \includegraphics[width=1\linewidth]{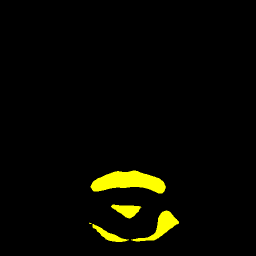}
                \caption*{IoU: 49.1}
            \end{subfigure}
            \begin{subfigure}[t]{0.32\textwidth}
                \centering
                \includegraphics[width=1\linewidth]{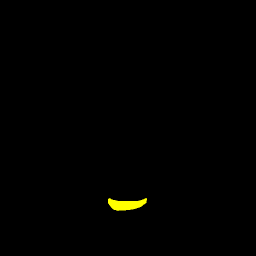}
                \caption*{IoU: 56.99}
            \end{subfigure}
        \end{subfigure}
        \vspace{-7pt}
        \caption{Predictions}    
    \end{subfigure}
    \caption{Example images in CelebA-HQ (a) with hand-annotations (b) and predictions (c) produced by the proposed model. Second row shows the 3 manual annotations of \emph{facial hair} label colored with red and 3 manual annotations of \emph{five o'clock shadow} colored with green. IoUs are given for the predictions with the corresponding hand-annotation.}
    \label{fig:celebA-HQ}
\end{figure*}

Previous works analyze the effect of facial hair on face recognition systems using a binary label,  categorizing images as clean-shaven or facial hair. We believe the degree of similarity or difference in facial hairstyles can have a significant effect on the match score for a pair of images. In this work, we present the first experimental analysis to understand the effect of facial hair size on similarity scores. First, a semantic segmentation model is trained to label facial hair pixels. CelebA-HQ \cite{celebahq} and MORPH \cite{morph} datasets are used to create 2550 manual annotations of the extent of facial hair. A cross-dataset evaluation is conducted, also comparing accuracy across demographic groups. Then, pre-trained ArcFace \cite{arcface} and AdaFace \cite{adaface} models are used to extract features and calculate similarity scores for different groups with varying extent of facial hair. We show that a large FMR variation can be observed on image pairs defined by facial hairstyle and this effect can be significantly reduced by employing adaptive thresholding.

This paper is organized as follows. In Section \ref{sec:related_work}, we review related work on facial hair segmentation and its impact on face recognition. In Section \ref{sec:facial_hair_segmentation}, we describe our dataset and experimental setup for building the segmentation model. Section \ref{sec:face_matching} introduces the face matchers and dataset we use to analyze the effects of facial hair on similarity scores. Then, findings on FMR variation and proposed thresohlding method based on facial hair is presented. Finally, in Section \ref{sec:conclusions}, we conclude with a summary and suggestions for future research.

\section{Related Work}
\label{sec:related_work}

\paragraph{Segmentation of facial hair region.}
In the literature, ``hair segmentation'' typically refers to hair growing from the scalp (head) and does not include facial hair \cite{shen2014image, svanera2016figaro, muhammad2018hair_figaro2, guo2016hair, levinshtein2018real, yoon2021real, yan2020two}.  Early work on facial hair segmentation by Ngyuen \etal \cite{nguyen2008image} proposes a layer extraction method to automatically decompose facial attributes such as beard and glasses. After decomposition, they apply post-processing to finalize the segmentation. They provide visual examples of segmentation results, but no quantitative results.

In \cite{le2012beard}, several steps are applied to detect beard and mustache regions. First, 79 facial landmarks are obtained using a modified Active Shape Model \cite{seshadri2009robust}. Next, the Self-Quotient \cite{wang2004face} algorithm is used to eliminate illumination artifacts. Then, four regions of interest (mustache, left beard, right beard and middle beard) are defined. A binary sparse classifier is used to label these  regions as skin or facial hair. In the follow-up work \cite{le2015fast}, the Self-Quotient step is removed to reduce the complexity. Instead, they build a feature vector consisting of Histogram of Gabor (HoG) and Histogram of Oriented Gradient of Gabor (HOGG) at different directions and frequencies, to detect facial hair on image patches. They report that this method is faster but less accurate than their previous work. These two approaches are later combined with the SLIC \cite{achanta2012slic} superpixel algorithm in \cite{le2017semi}. Neither the annotations nor the segmentation model from this work have been made publicly-available.  Also, the accuracy figures are for coarse subregions of the face.  For example, if an image is marked as hair in the left beard area of the image and any hair is found in the left beard area, it is counted as a binary match.  The spatial extent of the match is not considered.

\paragraph{Impact of facial hair on recognition accuracy.}
In an earlier work \cite{givens2004features} before the deep learning era, effects of facial hair with several facial attributes are analyzed. Three well-known algorithms with cosine similarity are applied to measure the distance between 2,144 face images from FERET \cite{phillips1998feret}. They only use a binary label to mark images as facial hair or no facial hair. Their results suggest that when facial hair is present in one image and not another, face recognition accuracy improves.

In \cite{lu2019experimental}, effects of 7 covariates, including facial hair, are analyzed using 5 DNNs. Four binary labels are used for facial hair: no facial hair, mustache, goatee and beard. They report that state-of-the-art (SOTA) deep models are able to handle facial hair variations, and facial hair does not change the key features of faces.

In \cite{terhorst2021comprehensive}, two popular deep learning models are employed to investigate the influence of facial hair on face verification. ``No beard'' and ``5 o'clock shadow'' binary attributes are compared, and they report that ``5 o'clock shadow'' results in much higher recognition rates.

The effects of scalp hair and facial hair on face recognition are investigated in \cite{bhatta2023gender}. While a segmentation network is utilized to segment scalp hair, they use the combination of Microsoft Face API \cite{MicrosoftFaceAPI} and Amazon Rekognition \cite{Amazonrekognition} to make a binary classification of images as clean-shaven or facial hair. They report that the accuracy of classifying clean-shaven is lower for African-American than for Caucasian, and suggest that a better algorithm is needed to detect facial hair accurately across demographic groups. 

In work similar to ours \cite{haiyu}, a facial hair attribute dataset is built and a facial hair classifier is trained to explore the effect of facial hair. They use binary labels to define beard area as \emph{clean-shaven}, \emph{chin area} and \emph{side to side} and do not take into account the specific extent of beard region. 

\paragraph{Novelty of this work.}
Previous works studying how facial hair impacts recognition accuracy only considered binary facial hair attributes such as beard/no-beard, whereas we propose to segment the facial hair region in each image to characterize facial hairstyle similarity of a pair.
No previous work has examined the accuracy of facial hair segmentation across demographic groups, whereas we compare accuracy of our segmenter across Caucasian and African-American.
We present the first experimental results to show that an adaptive thresholding scheme can reduce the FMR variation across categories of hairstyle-defined images pairs.

\section{Facial Hair Segmentation}
\label{sec:facial_hair_segmentation}

\subsection{Training Segmentation Model}

The Bilateral Segmentation Network (BiSeNet) \cite{bisenet} is used to train our facial hair segmenter. This architecture is designed to encode spatial information through Spatial Path and Context Path to provide sufficient receptive field. 
We did not observe accuracy improvements with Spatial Path, so we only use Context Path. 
The implementation \cite{bisenet_github} is trained to classify each pixel as \emph{facial hair} or \emph{not facial hair}. Accuracy is measured on two datasets as Intersection over Union (IoU) of the detected facial hair region with the manually specified facial hair region (``ground truth'').

CelebA-HQ \cite{celebahq} is used to create facial hair annotations in $1024\times1024$ resolution.
CelebA-HQ images correspond to a subset of CelebA \cite{celeba}, with steps applied to obtain consistent quality and center the images on the facial region.
These steps include: artifact removal, 4x super-resolution, mirror padding, Gaussian filtering, cropping and resampling to $1024\times1024$.

\begin{table*}[htb]
\begin{center}
\begin{tabular}{|c|c|c|}
\hline
Test Set & Training Size & IOU Facial Hair \\
\hline
\multirow{4}{*}{CelebA-HQ facial hair} & 500 & $82.93\pm1.2$ \\
& 1000 & $84.07\pm0.95$ \\
& 1500 & $84.61\pm0.73$ \\
& 2000 & $84.85\pm0.7$ \\
\hline
\multirow{4}{*}{CelebA-HQ five o'clock shadow} & 500 & $64.23\pm3.44$ \\
& 1000 & $64.74\pm1.43$ \\
& 1500 & $64.62\pm1.81$ \\
& 2000 & $64.3\pm2.68$ \\
\hline
\multirow{4}{*}{MORPH-AAM} & 500 & $74.13\pm4.01$ \\
& 1000 & $77.23\pm1.61$ \\
& 1500 & $78.63\pm0.93$ \\
& 2000 & $79.1\pm0.15$ \\
\hline
\multirow{4}{*}{MORPH-CM} & 500 & $77.06\pm2.16$ \\
& 1000 & $79.59\pm1.4$ \\
& 1500 & $81.15\pm1.02$ \\
& 2000 & $81.77\pm0.85$ \\
\hline
\hline
MORPH GT-1 vs GT-2 &  & $78.39$ \\
\hline
\end{tabular}
\end{center}
\caption{Accuracy on test sets. IoU for \emph{facial hair} label is reported on CelebA-HQ facial hair (150 \emph{facial hair} images), CelebA-HQ five o'clock shadow (50 \emph{five o'clock shadow} images), MORPH-AAM (100 African-American male images) and MORPH-CM (100 Caucasian male images). IoU between ground truths from  two human annotators is also reported in last row for the 200 MORPH images.}
\label{tab:iou_results}
\end{table*}

A total of 2550 images were manually annotated (using labelme \cite{labelme}) to outline the extent of the facial hair region. To establish the boundary between scalp hair and facial hair, facial hair above the bottom of the ear was not annotated. During  annotation, we observe that it is difficult to annotate a region consistently as \emph{facial hair} or not if facial hair is very short. Consequently, we expect noisier annotations for short facial hair. Poor lighting and low-quality images also result in noisier annotations. To minimize the effect of noisy labels on accuracy evaluation, we use two labels to mark the images: \emph{facial hair} and \emph{five o'clock shadow}. The remaining areas of the images are considered as \emph{not facial hair}. This allows us to divide our data into 3 parts: (1) images with only the \emph{facial hair} label, (2) images with both \emph{facial hair} and \emph{five o'clock shadow} labels, and (3) images with only \emph{five o'clock shadow} label. Our training set consists of images from all three subsets. However, the validation and test sets do not include images with \emph{five o'clock shadow}. We  create a second test set with only \emph{five o'clock shadow} to report accuracy separately for this.

CelebA-HQ images with facial hair annotations are used for training (2000 images), validation (150 images) and test (200 images) sets. Additionally, 1500 clean-shaven images in CelebA-HQ, including both male and female identities, are also used during training to learn clean-shaven faces. CelebA-HQ images and annotations are downsized to $512\times512$ as we did not observe improvement in accuracy with higher resolution. Scaling (0.75, 1.25, 1.5), horizontal flipping, rotation ($90^{\circ}$, $270^{\circ}$) and color jitter are used to augment the training data. $480\times480$ random crops are then obtained to feed the network. Additionally, we apply random JPEG compression as we observe improvements on MORPH images with this additional mode of augmentation. 

\subsection{Results}

The MORPH dataset \cite{morph} was created to support research on face aging and it has also become heavily used in studying demographic variation in accuracy \cite{Albiero2020training, albiero2021gendered, bhatta2023gender, haiyu, albiero2020bmvc}. For a cross-dataset evaluation, 100 images of African-American males (AAM) and 100 images of Caucasian males (CM) are annotated in $480\times400$ resolution. We use a facial hair attribute classifier \cite{haiyu} to select 10 images in each of 10 categories, for each of African-American males and Caucasian males in MORPH (200 images total). Facial hair categories \cite{haiyu} include \emph{chinArea short mustacheConnected}, \emph{chinArea medium mustacheConnected}, \emph{chinArea short noMustache}, \emph{chinArea short isolated}, \emph{sideToSide short mustacheConnected}, \emph{sideToSide medium mustacheConnected}, \emph{sideToSide short isolated}, \emph{sideToSide long mustacheConnected}, \emph{mustacheOnly long} and \emph{mustacheOnly short}. Annotations were made independently by two annotators, in order to be able to report the IoU between two independent manual annotations (see Table \ref{tab:iou_results}).

Accuracy is reported on two datasets. Note that we expect lower IoU accuracy on MORPH images because training and validation sets consist only of CelebA-HQ images. Also, image quality of MORPH images is lower than CelebA-HQ (our CelebA-HQ annotations include finer details of facial hair; however, it is not easy to capture details on MORPH, especially if facial hair is short, because of image compression). Since the predictions on five o'clock shadow are not as consistent as on longer facial hair, we divide the CelebA-HQ test set into 150 images labeled with only \emph{facial hair} label and another 50 images with only \emph{five o'clock shadow} label. Figure \ref{fig:celebA-HQ} shows examples of CelebA-HQ images with manual annotations and model predictions. 

Table \ref{tab:iou_results} lists mean and standard deviation of IoU accuracy for the 5 models trained and tested with 5 different data splits. To analyze the impact of training set size, we also measured the performance of networks trained with 500, 1000, and 1500 facial hair images. These subsets were randomly selected 3 times from a total of 2000 facial hair images for each of the 5 data splits. The model achieves $84.85\%$ IoU on the \emph{CelebA-HQ facial hair} set which consists of 150 images with only \emph{facial hair} label. IoU decreases to $64.3\%$ for \emph{CelebA-HQ five o'clock shadow} which is composed of 50 images annotated with \emph{five o'clock shadow} label. The reason for this drop in accuracy is the ambiguity of annotating near clean-shaven facial hair length as \emph{facial hair} or \emph{not facial hair} (Figure \ref{fig:celebA-HQ}). Note that IoU is reported for \emph{facial hair} in Table \ref{tab:iou_results}. IoU of \emph{not facial hair} label is measured as $98.74\%$ for test. This reflects the fact that higher IoU is generally expected on matching larger regions, as the disagreements on the boundary become a smaller fraction of the total area.  We didn't observe mislabelling for our test set of 200 clean-shaven images. 

\begin{table}
\begin{tabular}{|c|c|}
\hline
Facial Hair Ratio & IOU \\
\hline
$> 0$ \& $< 0.05$ & 71.24 \\
$\ge 0.05$ \& $< 0.1$ & 80.92 \\
$\ge 0.1$ \& $< 0.15$ & 86.01 \\
$\ge 0.15$ & 90.21 \\
\hline
\end{tabular}
\caption{IoU on CelebA-HQ facial hair set for 4 facial hair ratio ranges. Facial hair ratio equals to number of pixels annotated as facial hair / size of image}
\label{tab:iou_facial_hair_ratio_ranges}
\end{table}

\begin{figure*}[htb]
    \captionsetup[subfigure]{aboveskip=2pt,belowskip=5pt}
        \begin{subfigure}[t]{0.12\textwidth}
            \centering
            \includegraphics[width=\linewidth]{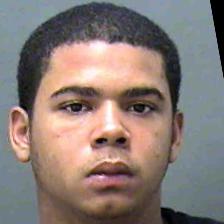}
        \end{subfigure}
        \begin{subfigure}[t]{0.12\textwidth}
            \centering
            \includegraphics[width=\linewidth]{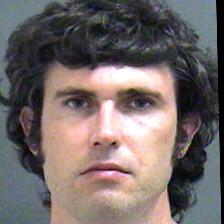}
        \end{subfigure}
        \begin{subfigure}[t]{0.12\textwidth}
            \centering
            \includegraphics[width=\linewidth]{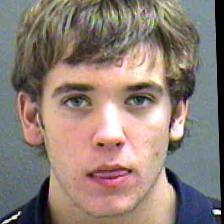}
        \end{subfigure}
        \begin{subfigure}[t]{0.12\textwidth}
            \centering
            \includegraphics[width=\linewidth]{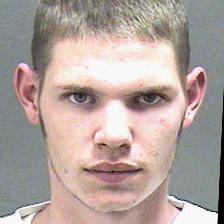}
        \end{subfigure}
        \begin{subfigure}[t]{0.12\textwidth}
            \centering
            \includegraphics[width=\linewidth]{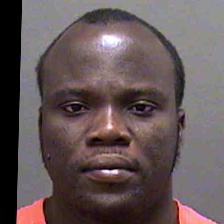}
        \end{subfigure}
        \begin{subfigure}[t]{0.12\textwidth}
            \centering
            \includegraphics[width=\linewidth]{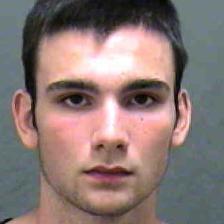}
        \end{subfigure}
        \begin{subfigure}[t]{0.12\textwidth}
            \centering
            \includegraphics[width=\linewidth]{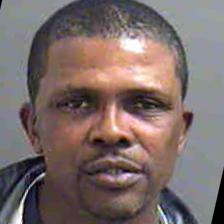}
        \end{subfigure}
        \begin{subfigure}[t]{0.12\textwidth}
            \centering
            \includegraphics[width=\linewidth]{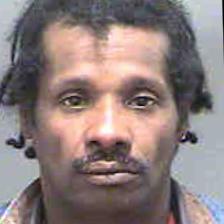}
        \end{subfigure}

        \begin{subfigure}[t]{0.12\textwidth}
            \centering
            \includegraphics[width=\linewidth]{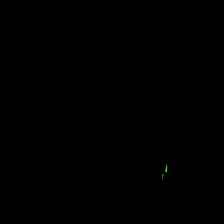}
            \caption*{0.00042}
        \end{subfigure}
        \begin{subfigure}[t]{0.12\textwidth}
            \centering
            \includegraphics[width=\linewidth]{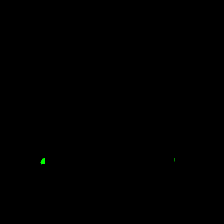}
            \caption*{0.00056}
        \end{subfigure}
        \begin{subfigure}[t]{0.12\textwidth}
            \centering
            \includegraphics[width=\linewidth]{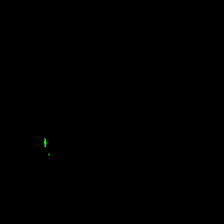}
            \caption*{0.00056}
        \end{subfigure}
        \begin{subfigure}[t]{0.12\textwidth}
            \centering
            \includegraphics[width=\linewidth]{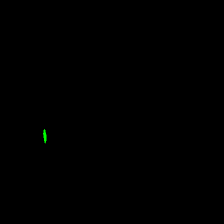}
            \caption*{0.00086}
        \end{subfigure}
        \begin{subfigure}[t]{0.12\textwidth}
            \centering
            \includegraphics[width=\linewidth]{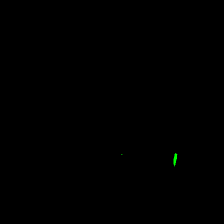}
            \caption*{0.00086}
        \end{subfigure}
        \begin{subfigure}[t]{0.12\textwidth}
            \centering
            \includegraphics[width=\linewidth]{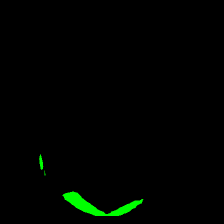}
            \caption*{0.01506}
        \end{subfigure}
        \begin{subfigure}[t]{0.12\textwidth}
            \centering
            \includegraphics[width=\linewidth]{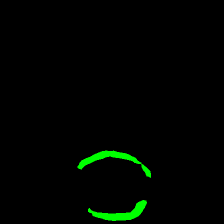}
            \caption*{0.01989}
        \end{subfigure}
        \begin{subfigure}[t]{0.12\textwidth}
            \centering
            \includegraphics[width=\linewidth]{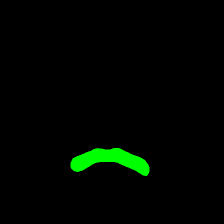}
            \caption*{0.02385}
        \end{subfigure}

        \begin{subfigure}[t]{0.12\textwidth}
            \centering
            \includegraphics[width=\linewidth]{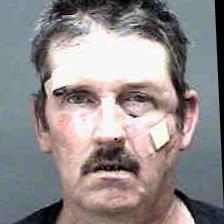}
        \end{subfigure}
        \begin{subfigure}[t]{0.12\textwidth}
            \centering
            \includegraphics[width=\linewidth]{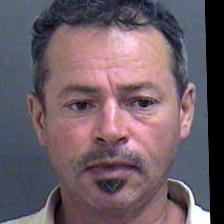}
        \end{subfigure}
        \begin{subfigure}[t]{0.12\textwidth}
            \centering
            \includegraphics[width=\linewidth]{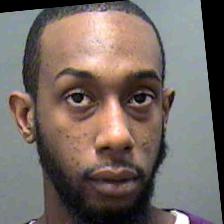}
        \end{subfigure}
        \begin{subfigure}[t]{0.12\textwidth}
            \centering
            \includegraphics[width=\linewidth]{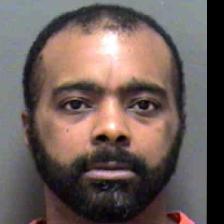}
        \end{subfigure}
        \begin{subfigure}[t]{0.12\textwidth}
            \centering
            \includegraphics[width=\linewidth]{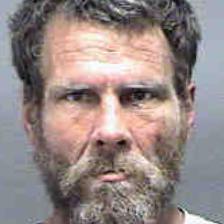}
        \end{subfigure}
        \begin{subfigure}[t]{0.12\textwidth}
            \centering
            \includegraphics[width=\linewidth]{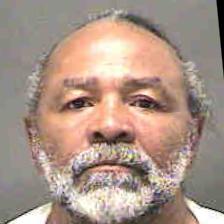}
        \end{subfigure}
        \begin{subfigure}[t]{0.12\textwidth}
            \centering
            \includegraphics[width=\linewidth]{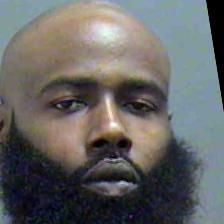}
        \end{subfigure}
        \begin{subfigure}[t]{0.12\textwidth}
            \centering
            \includegraphics[width=\linewidth]{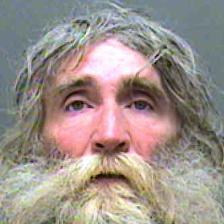}
        \end{subfigure}
        \begin{subfigure}[t]{0.12\textwidth}
            \centering
            \includegraphics[width=\linewidth]{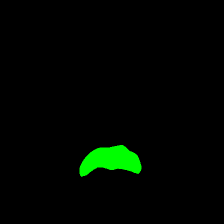}
            \caption*{0.02557}
        \end{subfigure}
        \begin{subfigure}[t]{0.12\textwidth}
            \centering
            \includegraphics[width=\linewidth]{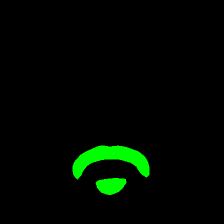}
            \caption*{0.14445}
        \end{subfigure}
        \begin{subfigure}[t]{0.12\textwidth}
            \centering
            \includegraphics[width=\linewidth]{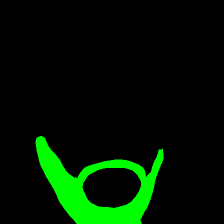}
            \caption*{0.14445}
        \end{subfigure}
        \begin{subfigure}[t]{0.12\textwidth}
            \centering
            \includegraphics[width=\linewidth]{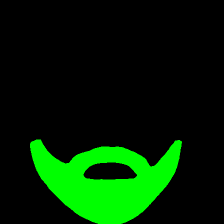}
            \caption*{0.16759}
        \end{subfigure}
        \begin{subfigure}[t]{0.12\textwidth}
            \centering
            \includegraphics[width=\linewidth]{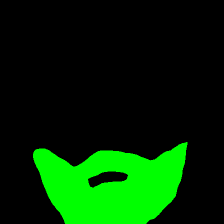}
            \caption*{0.17237}
        \end{subfigure}
        \begin{subfigure}[t]{0.12\textwidth}
            \centering
            \includegraphics[width=\linewidth]{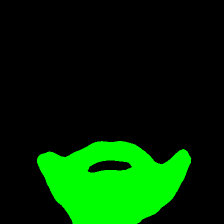}
            \caption*{0.17721}
        \end{subfigure}
        \begin{subfigure}[t]{0.12\textwidth}
            \centering
            \includegraphics[width=\linewidth]{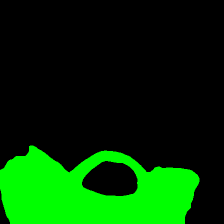}
            \caption*{0.21388}
        \end{subfigure}
        \begin{subfigure}[t]{0.12\textwidth}
            \centering
            \includegraphics[width=\linewidth]{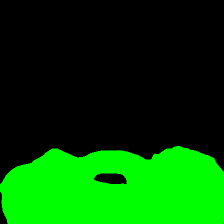}
            \caption*{0.29976}
        \end{subfigure}
    \caption{Morph images with varying facial hair ratios.}
    \label{fig:morph}
\end{figure*}

IoU accuracy on CelebA-HQ test images is shown in Table \ref{tab:iou_facial_hair_ratio_ranges} as a function of facial hair ratio. Facial hair ratio is the (number of pixels annotated as facial hair) / (size of image), ranging from 0 to about 35\% (see Figure \ref{fig:morph}). The same number of pixels incorrectly predicted as facial hair can be a large fraction of a small facial hair region or a small fraction of a large facial hair region. Thus, there is an overall trend of higher IoU accuracy for larger facial hair regions.

The results show that IoU on African-Americans ($79.1\%$) is slightly worse than Caucasians ($81.77\%$). One reason for this small difference is that, even though we use a facial attribute classifier \cite{haiyu} to balance the test sets for these 2 demographic groups, average facial hair ratios of ground truths is 0.063 for Caucasians and 0.054 for African-Americans. Note that, it is expected to observe lower IoU accuracy for smaller facial hair areas (see Table \ref{tab:iou_facial_hair_ratio_ranges}). In addition to the performance of the proposed model, IoU accuracy of two human annotators is also reported in Table \ref{tab:iou_results}. Surprisingly, annotations by two different persons only achieve $78.39\%$ IoU. This reflects the ambiguity of facial hair definition, especially in the presence of short facial hair, poor lighting and low quality images. 

\section{Face Matching}
\label{sec:face_matching}

In this work, we use two face recognition models to observe if the impact of facial hair on verification is consistent for two well-known SOTA face recognition models. ArcFace \cite{arcface} is one of the most popular models in recent years. A loss function ``Additive Angular Margin Loss'' is proposed to increase the discriminative power of features and help stabilize training. The version with ResNet100 \cite{resnet} backbone trained on Glint360K \cite{glint360k} is used in our experiments. AdaFace \cite{adaface}, which assigns different importance to samples based on image quality, is a more recent model that achieves SOTA accuracy on several datasets. It avoids focusing on images that are unidentifiable because of low quality, instead emphasizing hard-yet-recognizable instances.

\subsection{Impact of Facial Hair on Accuracy}
\label{sec:impact_of_facial_hair}

\begin{figure}[htb]
    \centering
    \includegraphics[width=\linewidth]{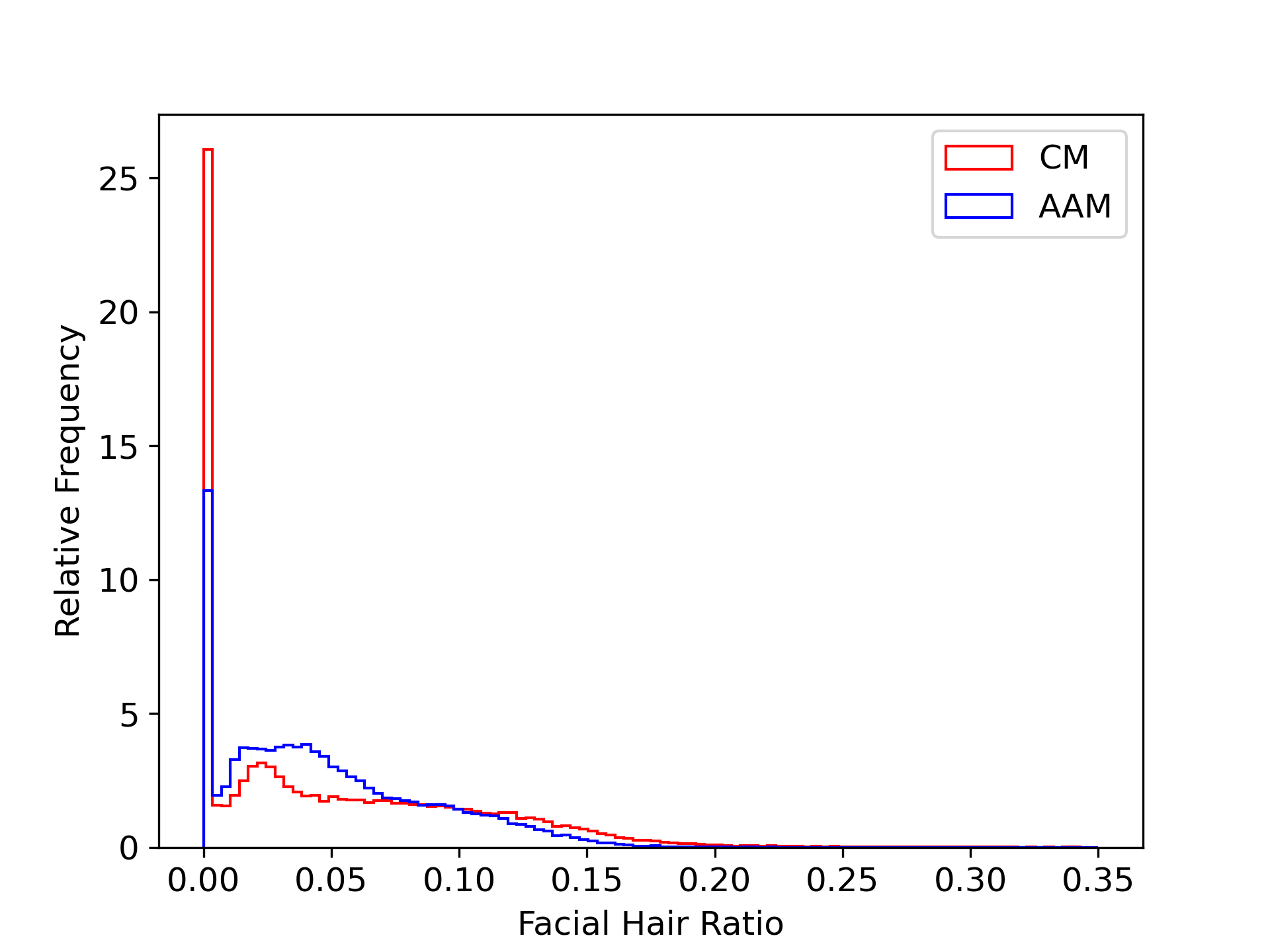}
    \caption{Distributions of facial hair ratio for African-American males and Caucasian males in MORPH. Spike at zero represents clean-shaven.}
    \label{fig:facial_hair_pred_hist}
\end{figure}

The curated version of MORPH in \cite{albiero2021gendered} is used to analyze the effect of facial hair size on face verification for African-American and Caucasian males. Face representations of 35,276 images of 8,835 Caucasian males and 56,245 images of 8,839 African-American males are obtained using the ArcFace and AdaFace models. Cosine similarity is used to obtain matching scores. Facial hair regions are detected by the segmentation model. Then, the number of \emph{facial hair} pixels is divided by the total number of pixels in the image to calculate facial hair ratio for creating pair groups. Figure \ref{fig:facial_hair_pred_hist} shows the distribution of facial hair ratio across African-American and Caucasian male images.

\begin{figure*}[htb]
    \centering
    \begin{subfigure}[t]{.75\textwidth}
        \centering
        \begin{subfigure}[t]{0.49\textwidth}
            \centering
            \includegraphics[width=0.95\linewidth]{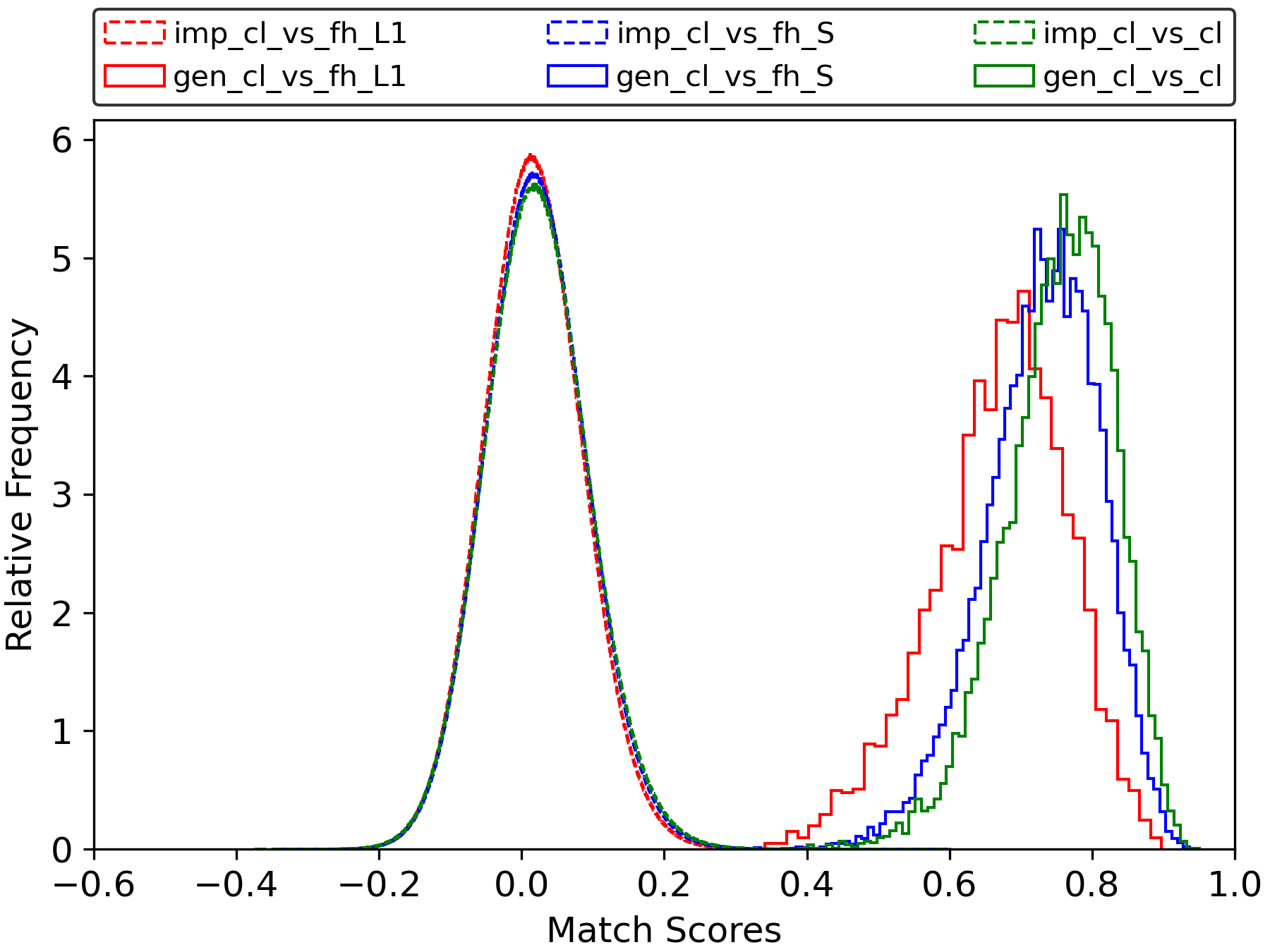}
            \label{fig:cl_vs_fh_CM}
        \end{subfigure}%
        \begin{subfigure}[t]{0.49\textwidth}
            \centering
            \includegraphics[width=0.95\linewidth]{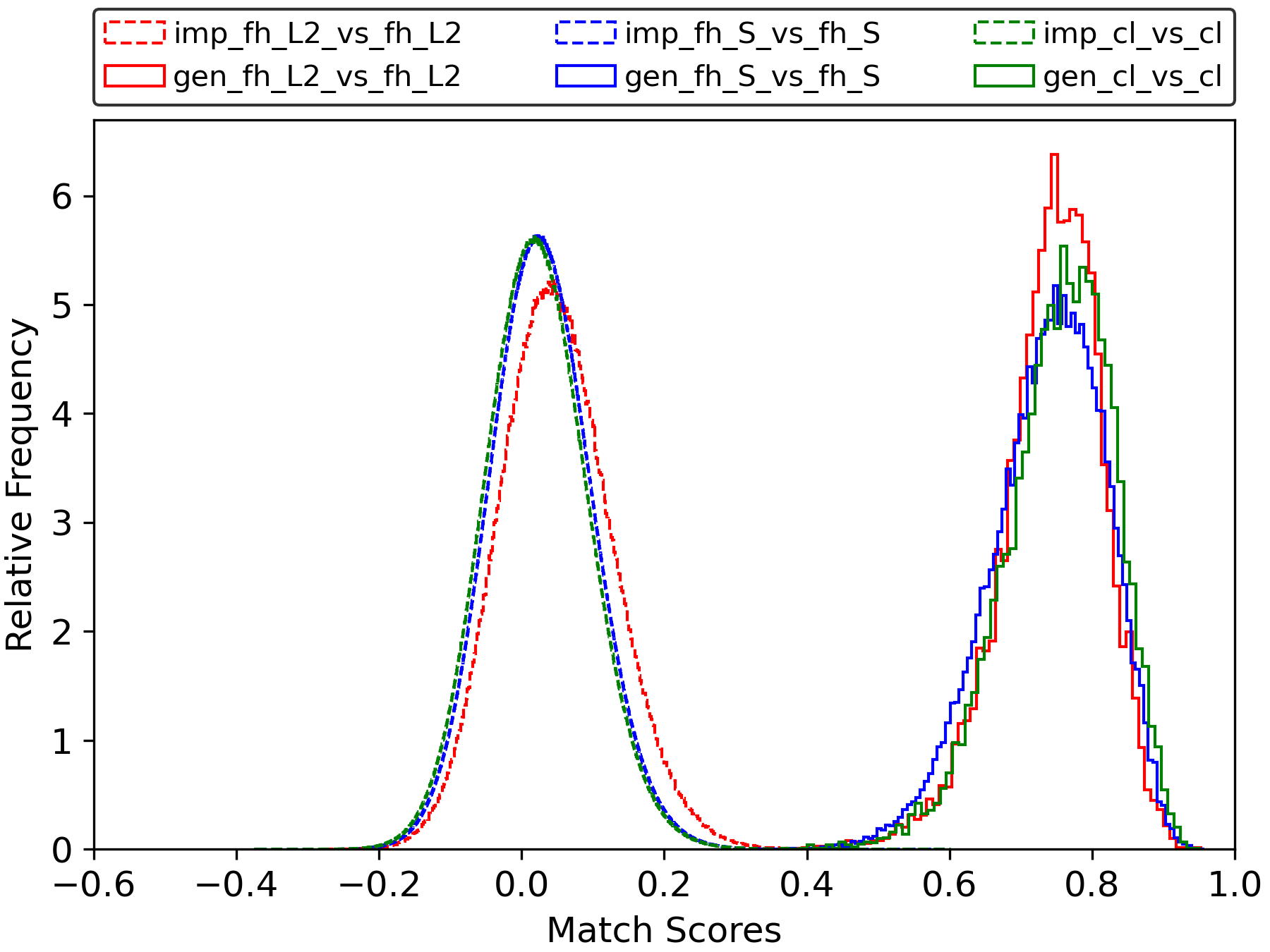}
            \hfill
            \label{fig:fh_vs_fh_CM}
        \end{subfigure}
        \caption{MORPH Caucasian Male (ArcFace)}
        \hfill
    \end{subfigure}
    \centering
    \begin{subfigure}[t]{.75\textwidth}
        \centering
        \begin{subfigure}[t]{0.49\textwidth}
            \centering
            \includegraphics[width=0.95\linewidth]{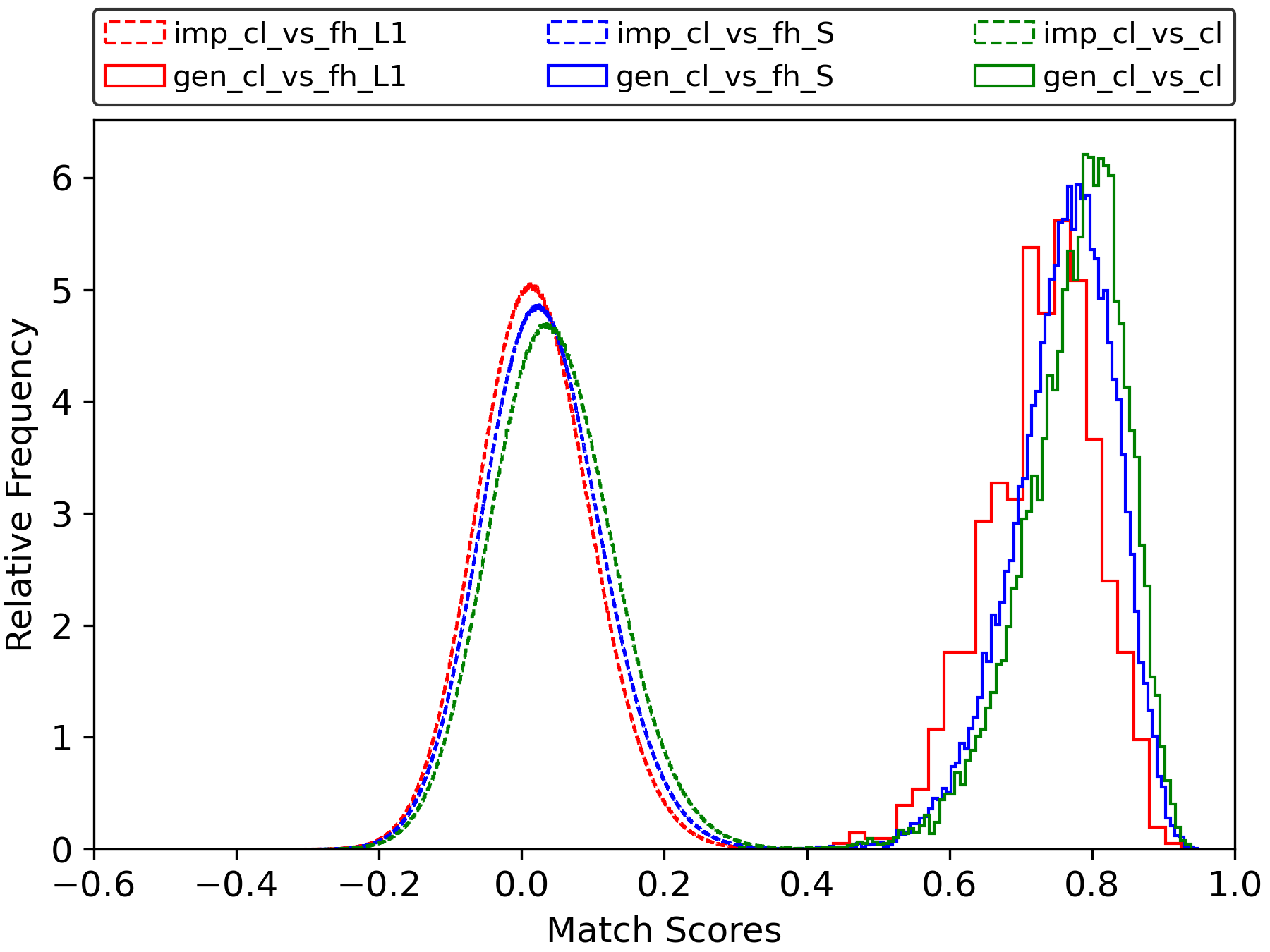}
            \label{fig:cl_vs_fh_AAM}
        \end{subfigure}%
        \begin{subfigure}[t]{0.49\textwidth}
            \centering
            \includegraphics[width=0.95\linewidth]{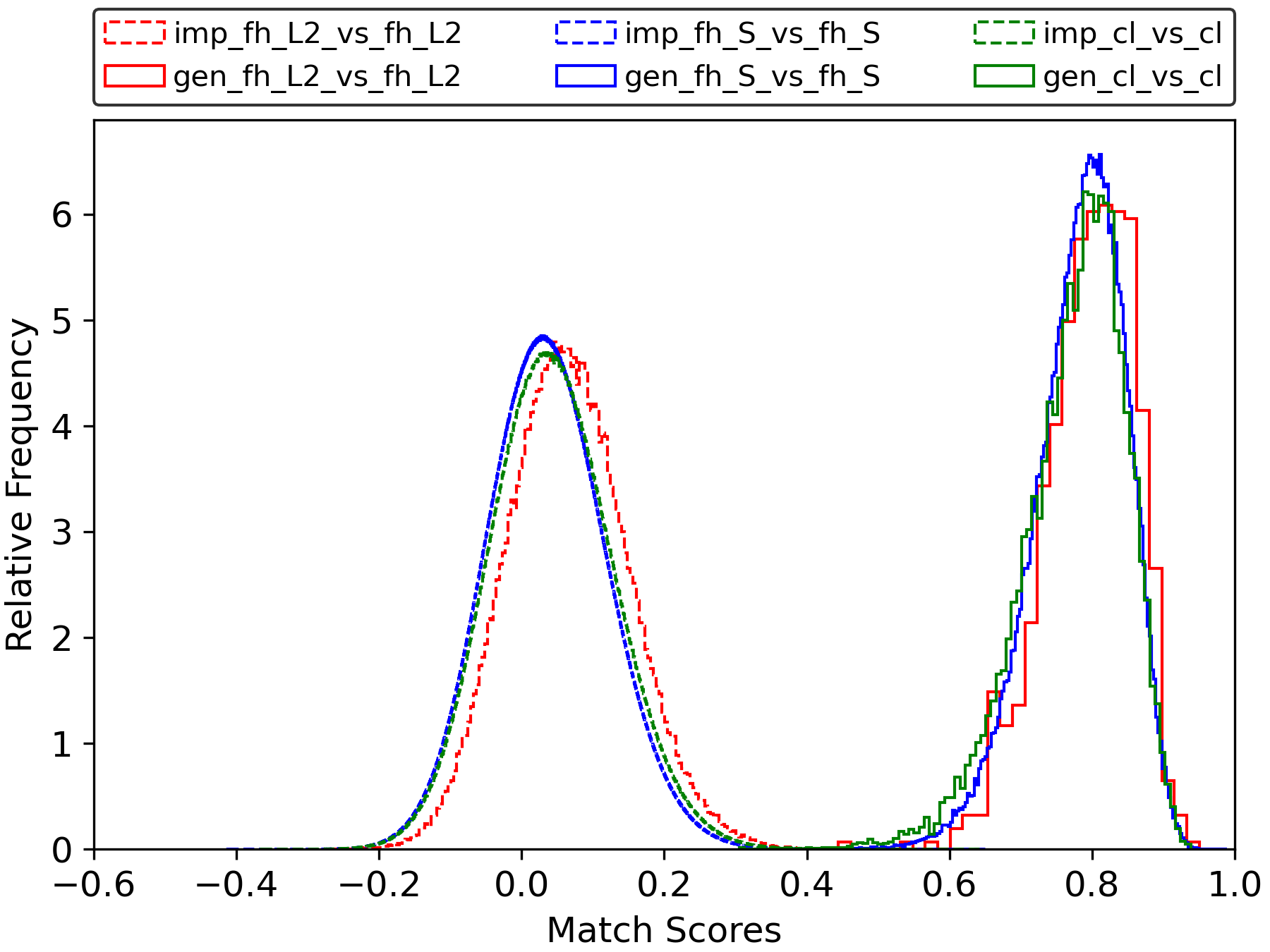}
            \hfill
            \label{fig:fh_vs_fh_AAM}
        \end{subfigure}
        \caption{MORPH African-American Male (ArcFace)}
    \end{subfigure}
    \centering
    \begin{subfigure}[t]{.75\textwidth}
        \centering
        \begin{subfigure}[t]{0.49\textwidth}
            \centering
            \includegraphics[width=0.95\linewidth]{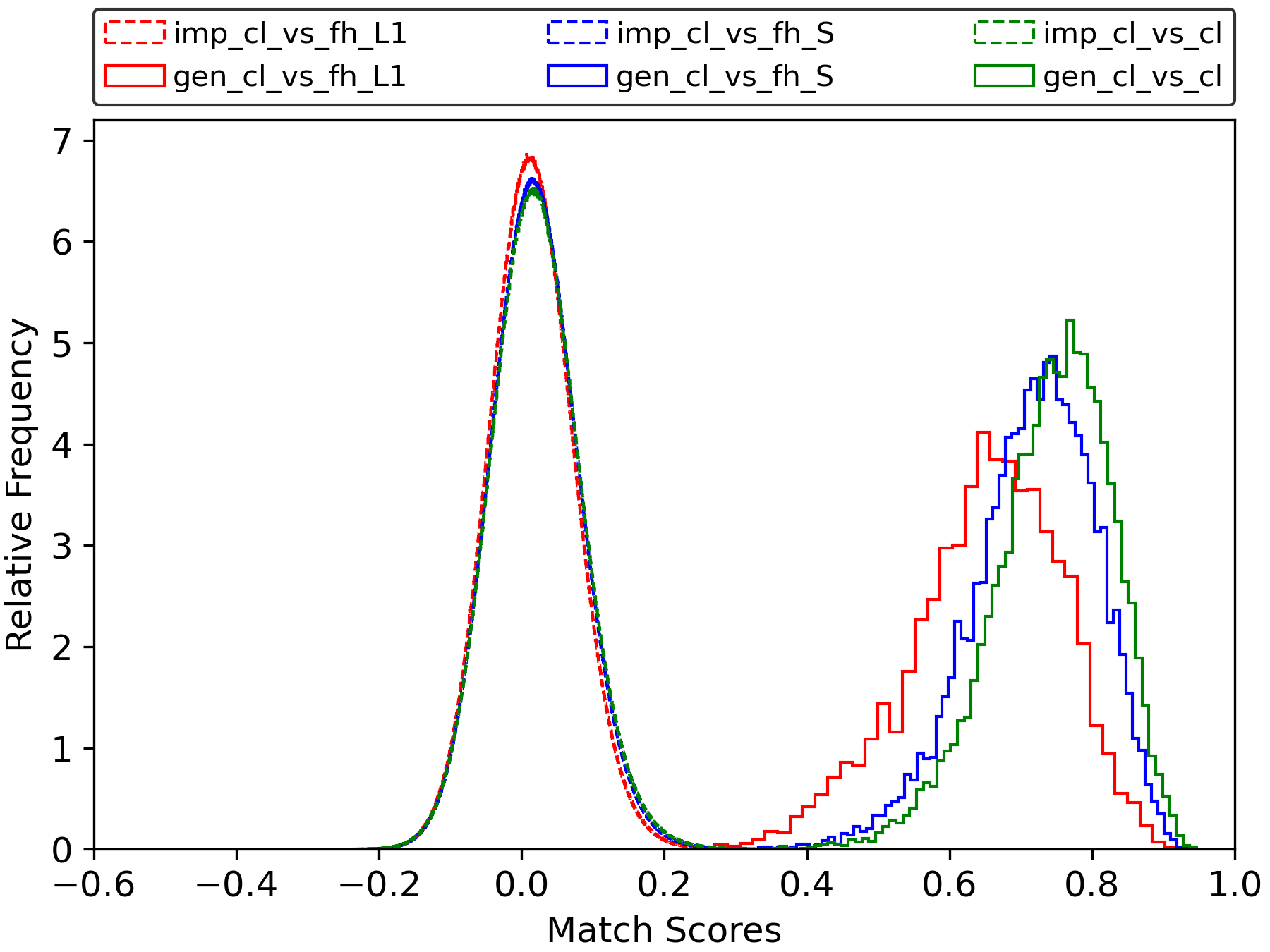}
            \label{fig:cl_vs_fh_CM_ada}
        \end{subfigure}%
        \begin{subfigure}[t]{0.49\textwidth}
            \centering
            \includegraphics[width=0.95\linewidth]{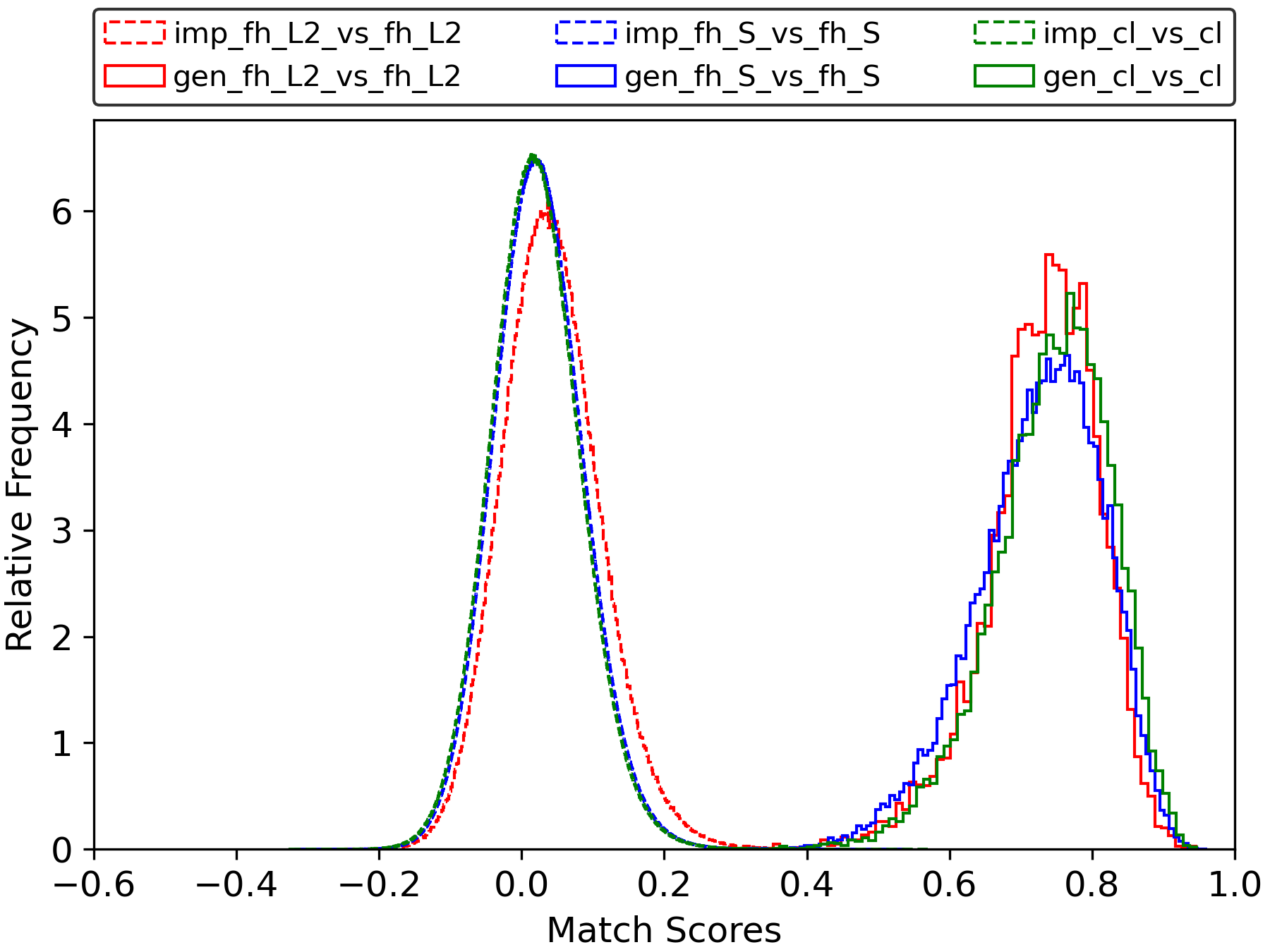}
            \hfill
            \label{fig:fh_vs_fh_CM_ada}
        \end{subfigure}
        \caption{MORPH Caucasian Male (AdaFace)}
    \end{subfigure}
    \centering
    \begin{subfigure}[t]{.75\textwidth}
        \centering
        \begin{subfigure}[t]{0.49\textwidth}
            \centering
            \includegraphics[width=0.95\linewidth]{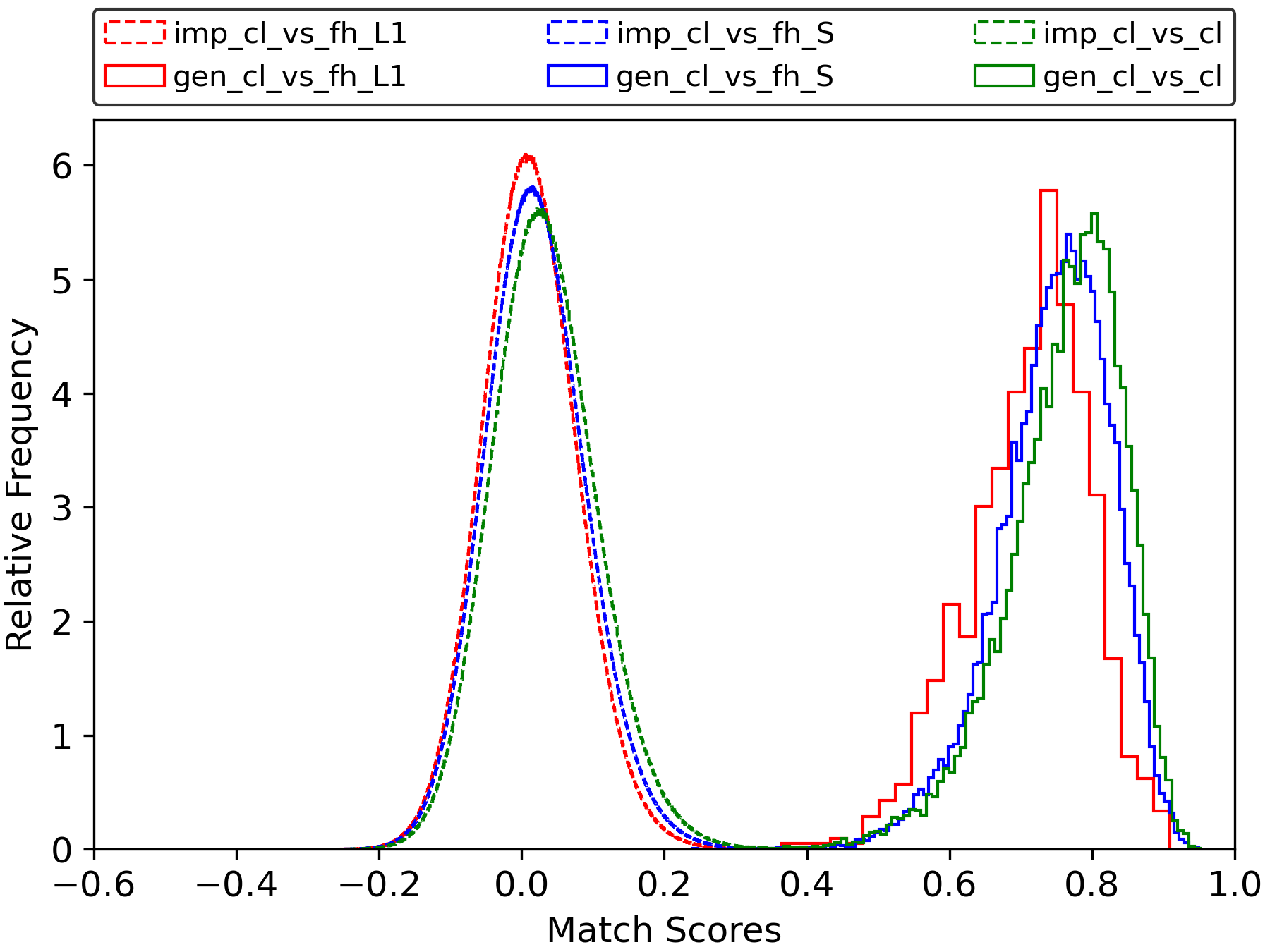}
            \label{fig:cl_vs_fh_AAM_ada}
        \end{subfigure}%
        \begin{subfigure}[t]{0.49\textwidth}
            \centering
            \includegraphics[width=0.95\linewidth]{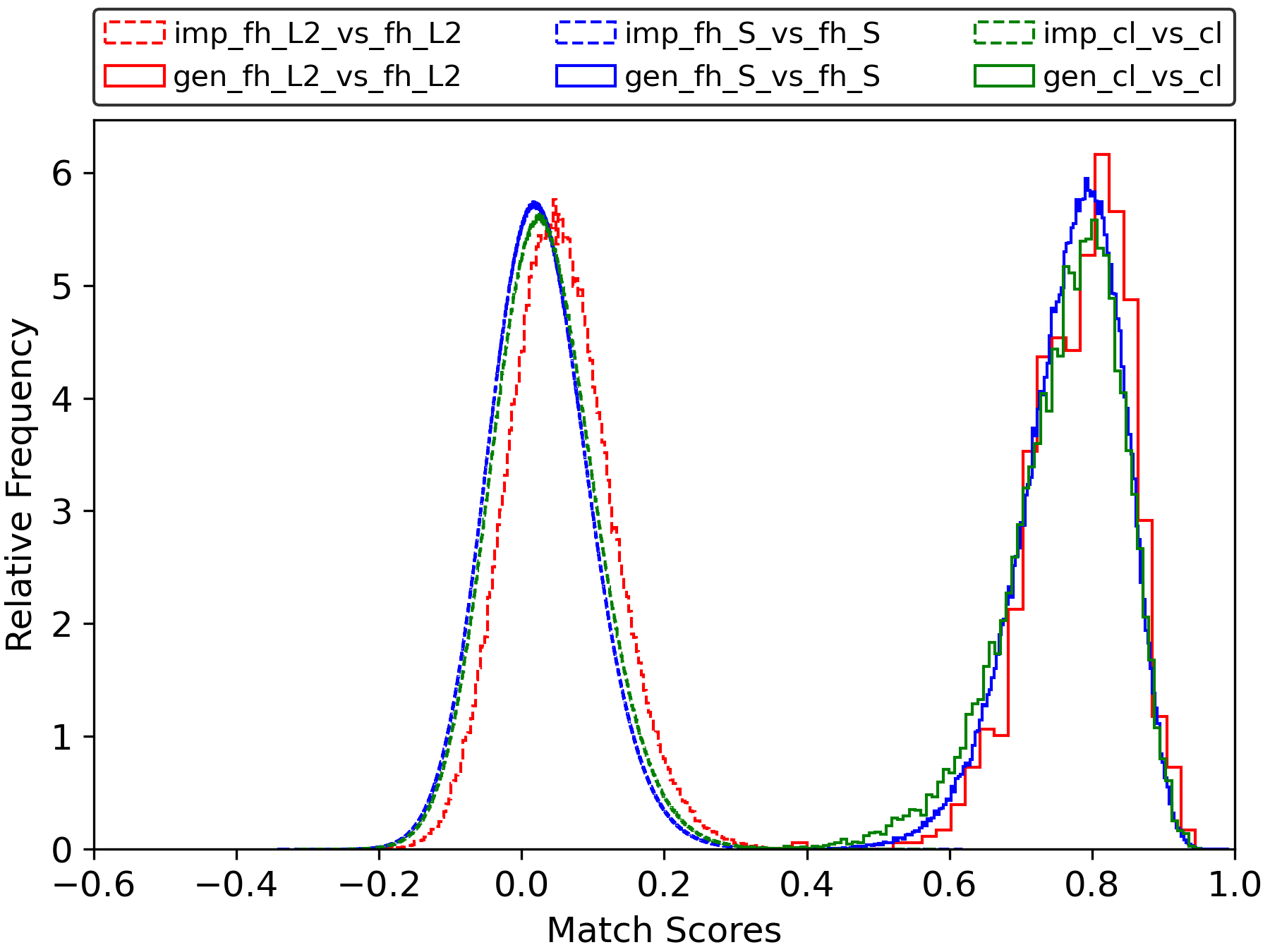}
            \hfill
            \label{fig:fh_vs_fh_AAM_ada}
        \end{subfigure}
        \caption{MORPH African-American Male (AdaFace)}
    \end{subfigure}
    \caption{Impostor and genuine similarity score distributions of Caucasians and African-Americans using ArcFace and AdaFace. Pair groups are built using 4 different image sets according to predicted facial hair ratio: \texttt{cl} is below 0.001, \texttt{fh\_S} is between 0.001 and 0.1, \texttt{fh\_L1} is above 0.1 and \texttt{fh\_L2} is above 0.15.}
    \label{fig:plots}
\end{figure*}

It is visually observed that near clean-shaven length hair can cause the segmentation network to predict small areas as \emph{facial hair}. Furthermore, since we determined not to label facial hair above the bottom of the ear, face images with sideburns may also lead to predictions of small regions. For this reason, we count the images as clean-shaven if the facial hair ratio is $< 0.001$. Two other thresholds, 0.1 and 0.15 are set to analyze the impact of the extent of the facial hair region. In our experiments, we build the following impostor and authentic pair groups: \emph{clean-shaven vs. clean-shaven} (\texttt{cl\_vs\_cl}), \emph{clean-shaven vs. facial hair} (\texttt{cl\_vs\_fh}) and \emph{facial hair vs. facial hair} (\texttt{fh\_vs\_fh}). Four facial hair ranges; ($< 0.001$), ($\ge 0.001$ \& $< 0.1$), ($\ge 0.1$) and ($\ge 0.15$) are depicted respectively as \texttt{cl}, \texttt{fh\_S}, \texttt{fh\_L1} and \texttt{fh\_L2} in the Figure \ref{fig:plots}.

\begin{figure*}[htb]
    \captionsetup[subfigure]{aboveskip=2pt,belowskip=5pt}
    \begin{subfigure}[t]{\textwidth}
        \centering
        \begin{subfigure}[t]{0.32\textwidth}
            \centering
            \includegraphics[width=\linewidth]{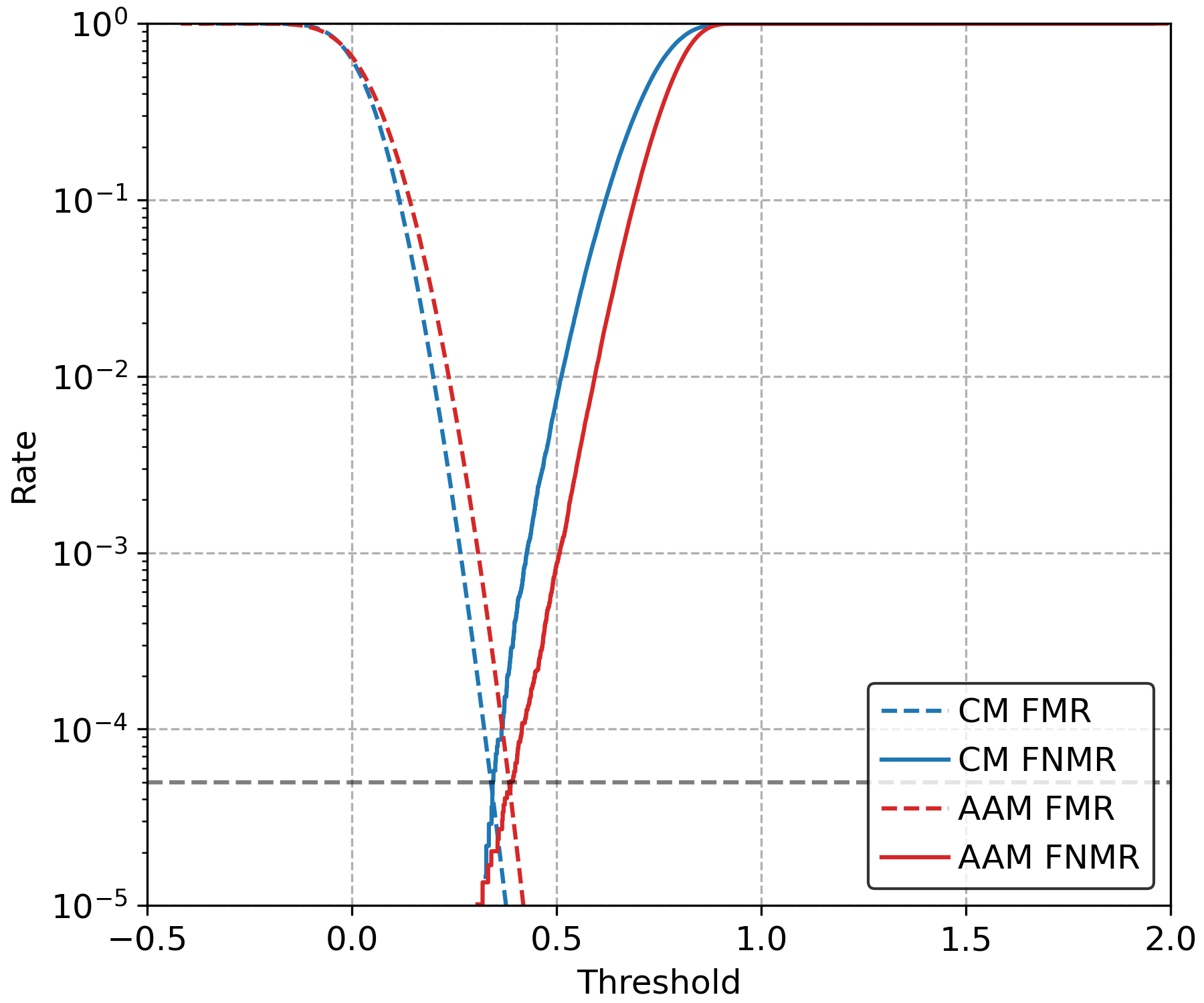}
            \caption{All pairs}
            \label{fig:fmr-all}
        \end{subfigure}
        \begin{subfigure}[t]{0.32\textwidth}
            \centering
            \includegraphics[width=\linewidth]{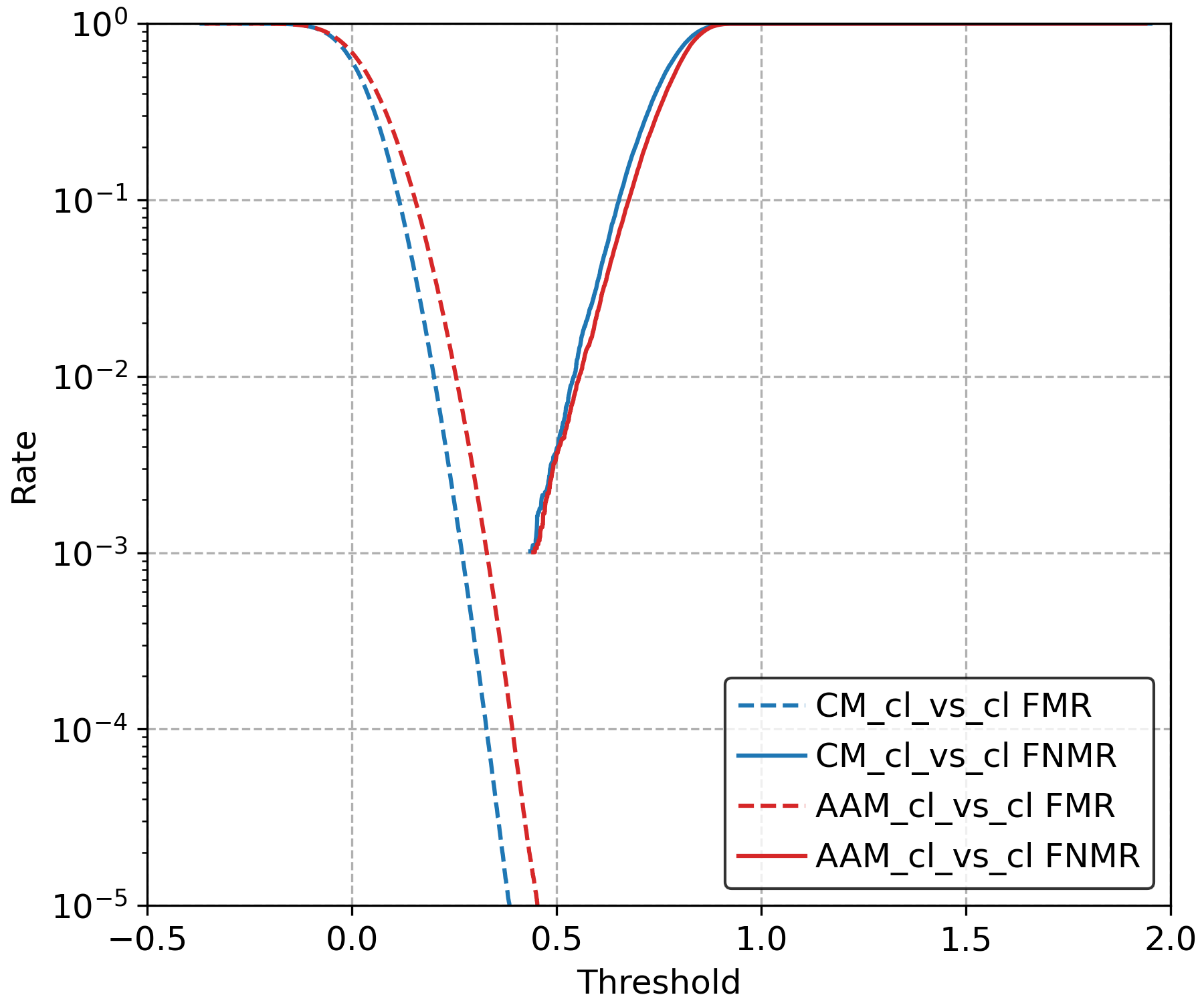}
            \caption{\texttt{cl\_vs\_cl}}
        \end{subfigure}
        \begin{subfigure}[t]{0.32\textwidth}
            \centering
            \includegraphics[width=\linewidth]{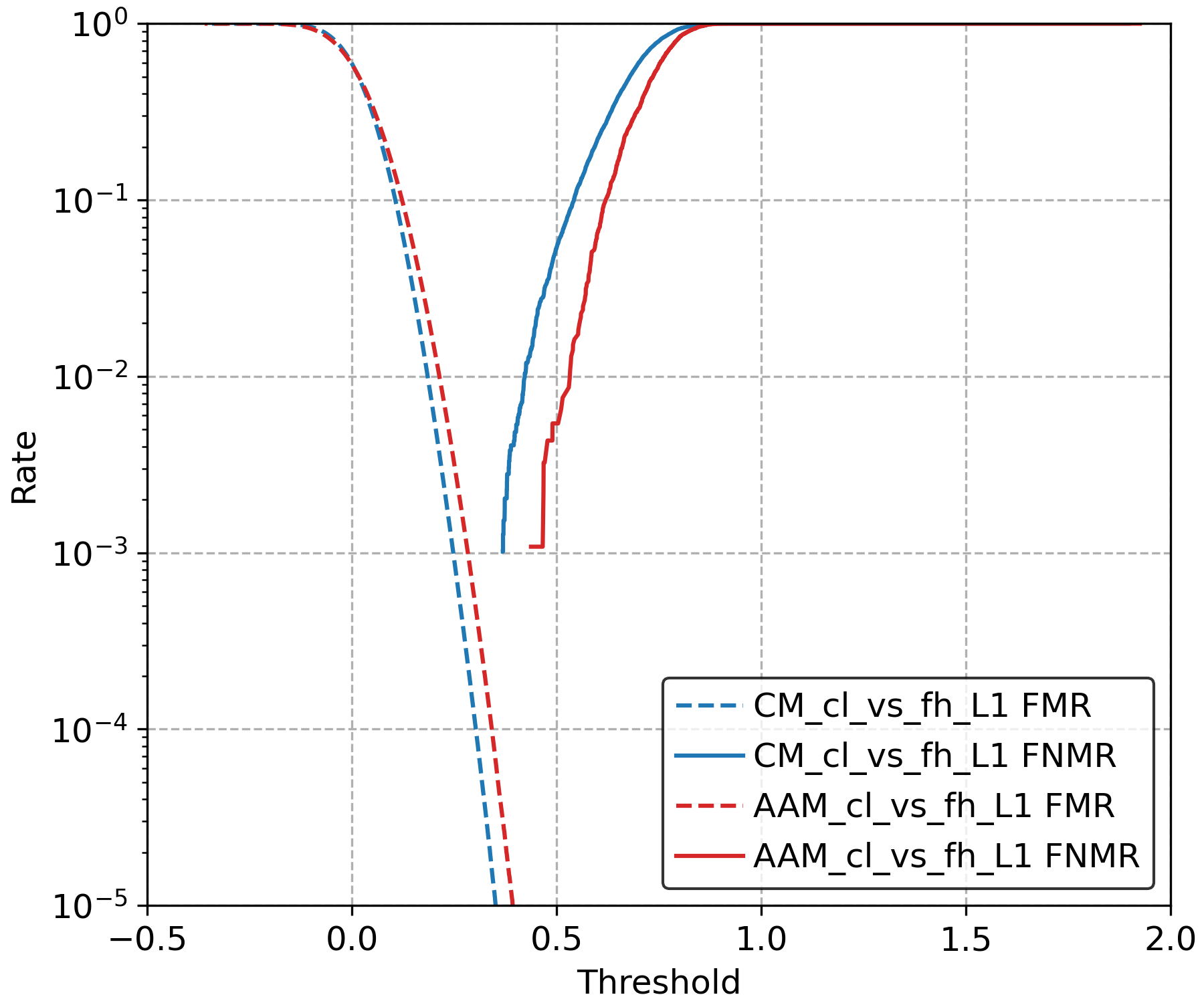}
            \caption{\texttt{cl\_vs\_fh\_L1}}
        \end{subfigure}
    \end{subfigure}
    \caption{Caucasian and African-American FMR and FNMR curves. (a) all pairs;  (b) \texttt{cl\_vs\_cl} subset; (c) \texttt{cl\_vs\_fh\_L1} subset.}
    \label{fig:fmr}
\end{figure*}

\begin{table*}[htb]
    \centering
    \begin{tabular}{|c|c|c|c||c|}
    \hline
    & cl vs. cl($10^{-4}$) & cl vs. fh\_L1($10^{-4}$) & fh\_L2 vs. fh\_L2($10^{-4}$) & max(FMR) / min(FMR)\\
    \hline
    
    AAM (Global Thr) & $2.55\pm0.09$ & $0.33\pm0.02$ & $3.61\pm0.58$ & $10.79\pm1.54$\\

    AAM (Adaptive Thr) & $0.97\pm0.12$ & $1.04\pm0.07$ & $0.96\pm0.82$ & $1.78\pm0.32$ (3 splits)\\
    
    \hline
    
    CM (Global Thr) & $1.28\pm0.04$ & $0.39\pm0.02$ & $10.01\pm0.54$ & $25.87\pm1.61$\\

    CM (Adaptive Thr) & $1.03\pm0.04$ & $1.01\pm0.05$ & $1.21\pm0.24$ & $1.27\pm0.23$\\
    \hline
    \end{tabular}
    \caption{FMR($10^{-4}$) on test set for three pair groups. Threshold values are determined from the validation set. Results are reported for 5 random splits (zero FMR is observed in two splits for African-Americans using adaptive threshold, thus max(FMR) / min(FMR) is reported using the other three data splits). Facial hair ratios used to create the pairs: \texttt{cl} is below 0.001, \texttt{fh\_L1} is above 0.1 and \texttt{fh\_L2} is above 0.15. The last column shows the ratio of maximum FMR to minimum FMR across pair groups.}
    \label{tab:adaptive_thr}
\end{table*}

Similarity score distributions of the different pair groups are shown in Figure \ref{fig:plots} for African-Americans and Caucasians. The pair group of \texttt{cl\_vs\_cl} is included in the plots in the first and second columns to act as a baseline for comparison. It can be observed by looking at the \texttt{cl\_vs\_fh\_L1}, \texttt{cl\_vs\_fh\_S} and \texttt{cl\_vs\_cl} pairs of both demographics that, the greater amount of facial hair size that differs between the images leads to lower similarity scores on genuine pairs (Figure \ref{fig:plots} left column). The same effect can be observed on impostor pairs of African-Americans. However, Caucasian impostor distributions are similar to each other. 

On the other hand, if both images in a genuine pair have a great amount of facial hair \texttt{gen\_fh\_L2\_vs\_fh\_L2}, similarity scores get closer to \texttt{gen\_cl\_vs\_cl} if it is not higher. However, \texttt{imp\_fh\_L2\_vs\_fh\_L2} impostor distributions of 2 demographic groups are shifted to the higher scores significantly compared to \texttt{imp\_cl\_vs\_cl} impostors. For these reasons, it can be claimed that adaptive thresholds can be used, according to the amount of facial hair, to improve verification performance. For example, if the reference image has a great amount of facial hair, we can expect lower scores for a clean-shaven query face image and much higher similarity scores for a comparison with an image that has a similar amount of facial hair as the reference image (\eg mean of \texttt{gen\_cl\_vs\_fh\_L1} is $0.67$ compared to mean of \texttt{gen\_fh\_L2\_vs\_fh\_L2} that is $0.74$ for Caucasians).

False Match Rate (FMR) and False Non-Match Rate (FNMR) analysis is employed across pair groups to assess the facial hair impact quantitatively. First, we find a threshold value to have a 1-in-10,000 FMR for Caucasians and African-Americans separately. We observe that, FMR of \texttt{imp\_fh\_L2\_vs\_fh\_L2} pairs increases to $4.23 \times 10^{-4}$ for African-Americans and $9.47 \times 10^{-4}$ for Caucasians. Note that, \texttt{imp\_cl\_vs\_fh\_L1} has the lowest FMR among our pair groups ($3.17 \times 10^{-5}$ for AAM and $3.9 \times 10^{-5}$ for CM).
An Inequity ratio of (max FMR across groups)/(min FMR across groups) proposed by the NIST FRVT report (see section 2.2 of \cite{Grother2022}) is used to measure differences between pair groups.
Using this measure for the facial hair categories indicates a hairstyle-based inequity of 24 for CM and 13 for AAM.
These results suggest that partitioning pairs into subgroups based on facial hair size can be useful to detect subsets with high FMR and help maintaining the same FMR rate across all groups by determining threshold values according to facial hair.

Disparity between the similarity scores of African-Americans and Caucasians is further investigated in Figure \ref{fig:fmr}. Since the plots obtained using AdaFace and ArcFace are quite similar, we only use the ArcFace model in Figure \ref{fig:fmr}. It can be seen that, both demographics reach an Equal Error Rate around $5 \times 10^{-5}$ indicating there is no significant gap in verification performance (see Figure \ref{fig:fmr-all}). Even though African-American pairs have higher FMR for all three groups, they have lower FNMR ((a) all pairs, (b) \texttt{cl\_vs\_cl} and (c) \texttt{cl\_vs\_fh\_L1} in Figure \ref{fig:fmr}). It is important to note that, their FNMR are getting closer to each other for \texttt{cl\_vs\_cl} and moving away from each other for \texttt{cl\_vs\_fh\_L1}. Since we have low error rates, a larger dataset with more genuine pairs is needed for a more accurate analysis.

\subsection{Adaptive Threshold to Mitigate Bias}

We conduct an experiment to show that facial hair predictions can be used to find adaptive thresholds for pair groups, created according to the size of facial hair, to alleviate the facial hair effect on targeted FMR. We first randomly divide the MORPH dataset into two parts: validation and test. Note that they have the same number of disjoint subjects, but they can have a different number of pairs according the number of clean-shaven and facial hair images. To have a 1-in-10,000 FMR, a global threshold is found by looking the all impostor similarity scores in the validation set. Also, threshold values for three subgroups, \texttt{cl\_vs\_cl}, \texttt{cl\_vs\_fh\_L1} and \texttt{fh\_L2\_vs\_fh\_L2} are determined to achieve our FMR target on each group. Then, these threshold values are used on the disjoint test set to report results. The experimental protocol is applied separately for African-Americans and Caucasians.

Table \ref{tab:adaptive_thr} shows the FMR on the test set using threshold values determined by the validation set. Mean and standard deviation of the 5 random splits are reported. Ratios of the worst and best FMR \cite{Grother2022} across facial hair groups is also given in the last column. Due to low number of impostor pairs, zero FMR is observed in two splits for African-Americans using adaptive threshold, thus mean and standard deviation is reported using the other three data splits. Results show that there can be a 10 times difference between the highest FMR and lowest FMR across groups based on the facial hair for African-Americans and 25 times difference for Caucasians (similar results are observed for the whole dataset before validation-test split, 13 for African-Americans and 24 for Caucasians. See Section \ref{sec:impact_of_facial_hair}). We show that this impact can be diminished using an adaptive threshold that is determining a threshold based on the facial hair size. The ratio between the worst and best FMR for African-American reduces from 10.70 to 1.78 and from 25.87 to 1.27 for Caucasians.

\section{Conclusions}
\label{sec:conclusions}

To investigate how facial hair impacts recognition accuracy, we first train a semantic segmentation model to segment facial hair.
IoU accuracy of the trained model is evaluated cross-dataset, and compared across African-American and Caucasian.
This is the only facial hair segmenter to be evaluated across demographics, and the implementation is made available to the research community.

Using facial hair segmentations from our model, we analyze the impostor and genuine distribution based on the extent of facial hair in the pair of images. We find that the greater the difference in facial hair size between two images in an impostor or genuine pair, the lower the similarity score.
Categories of facial hairstyles are defined for a pair of images, and large differences in FMR are observed between categories.

Our findings demonstrate that utilizing a variable threshold determined by the amount of facial hair present can substantially mitigate the FMR bias due to facial hairstyle.
The FMR max-min ratio for both African-American and Caucasian drops from nearly 10 and 25, respectively, to below 2 for both demographics.
The variable threshold also makes it harder for someone to use facial hair to change their appearance in order to create a false non-match result.

\clearpage
{\small
\bibliographystyle{ieee_fullname}
\bibliography{egbib}

\begin{thebibliography}{10}\itemsep=-1pt

\bibitem{abdurrahim2018review}
Salem~Hamed Abdurrahim, Salina~Abdul Samad, and Aqilah~Baseri Huddin.
\newblock Review on the effects of age, gender, and race demographics on
  automatic face recognition.
\newblock {\em The Vis. Comput.}, 34:1617--1630, 2018.

\bibitem{achanta2012slic}
Radhakrishna Achanta, Appu Shaji, Kevin Smith, Aurelien Lucchi, Pascal Fua, and
  Sabine S{\"u}sstrunk.
\newblock Slic superpixels compared to state-of-the-art superpixel methods.
\newblock {\em IEEE Transactions on Pattern Analysis and Machine Intelligence},
  34(11):2274--2282, 2012.

\bibitem{albiero2020bmvc}
V{\'\i}tor Albiero and Kevin~W Bowyer.
\newblock Is face recognition sexist? {N}o, gendered hairstyles and biology
  are.
\newblock In {\em Proceedings of the British Machine Vision Conference (BMVC)},
  2020.

\bibitem{Albiero2020training}
Vítor Albiero, Kai Zhang, and Kevin~W. Bowyer.
\newblock How does gender balance in training data affect face recognition
  accuracy?
\newblock In {\em 2020 IEEE International Joint Conference on Biometrics
  (IJCB)}, 2020.

\bibitem{albiero2021gendered}
V{\'\i}tor Albiero, Kai Zhang, Michael~C King, and Kevin~W Bowyer.
\newblock Gendered differences in face recognition accuracy explained by
  hairstyles, makeup, and facial morphology.
\newblock {\em IEEE Transactions on Information Forensics and Security},
  17:127--137, 2021.

\bibitem{Amazonrekognition}
\url{https://aws.amazon.com/rekognition/}.

\bibitem{glint360k}
Xiang An, Xuhan Zhu, Yuan Gao, Yang Xiao, Yongle Zhao, Ziyong Feng, Lan Wu, Bin
  Qin, Ming Zhang, Debing Zhang, et~al.
\newblock Partial fc: Training 10 million identities on a single machine.
\newblock In {\em Proceedings of the IEEE/CVF International Conference on
  Computer Vision}, pages 1445--1449, 2021.

\bibitem{bhatta2023gender}
Aman Bhatta, V{\'\i}tor Albiero, Kevin~W Bowyer, and Michael~C King.
\newblock The gender gap in face recognition accuracy is a hairy problem.
\newblock In {\em Proceedings of the IEEE/CVF Winter Conference on Applications
  of Computer Vision}, pages 303--312, 2023.

\bibitem{bisenet_github}
\url{https://github.com/zllrunning/face-parsing.PyTorch}.

\bibitem{9209125}
Jacqueline~G. Cavazos, P.~Jonathon Phillips, Carlos~D. Castillo, and Alice~J.
  O’Toole.
\newblock Accuracy comparison across face recognition algorithms: Where are we
  on measuring race bias?
\newblock {\em IEEE Transactions on Biometrics, Behavior, and Identity
  Science}, 3(1):101--111, 2021.

\bibitem{arcface}
Jiankang Deng, Jia Guo, Jing Yang, Niannan Xue, Irene Kotsia, and Stefanos
  Zafeiriou.
\newblock Arcface: Additive angular margin loss for deep face recognition.
\newblock {\em IEEE Transactions on Pattern Analysis and Machine Intelligence},
  44(10):5962--5979, 2021.

\bibitem{9086771}
Pawel Drozdowski, Christian Rathgeb, Antitza Dantcheva, Naser Damer, and
  Christoph Busch.
\newblock Demographic bias in biometrics: A survey on an emerging challenge.
\newblock {\em IEEE Transactions on Technology and Society}, 1(2):89--103,
  2020.

\bibitem{georgopoulos2021mitigating}
Markos Georgopoulos, James Oldfield, Mihalis~A Nicolaou, Yannis Panagakis, and
  Maja Pantic.
\newblock Mitigating demographic bias in facial datasets with style-based
  multi-attribute transfer.
\newblock {\em International Journal of Computer Vision}, 129(7):2288--2307,
  2021.

\bibitem{givens2004features}
G Givens, JR Beveridge, BA Draper, P Grother, and PJ Phillips.
\newblock How features of the human face affect recognition: a statistical
  comparison of three face recognition algorithms.
\newblock In {\em Proceedings of the 2004 IEEE Computer Society Conference on
  Computer Vision and Pattern Recognition}, pages 381--389, 2004.

\bibitem{Grother2022}
Patrick Grother.
\newblock Face recognition vendor test ({FRVT}) part 8: Summarizing demographic
  differentials.
\newblock Technical report, 2022.

\bibitem{grother2019face}
Patrick Grother, Mei Ngan, and Kayee Hanaoka.
\newblock {\em Face recognition vendor test (fvrt): Part 3, demographic
  effects}.
\newblock National Institute of Standards and Technology Gaithersburg, MD,
  2019.

\bibitem{guo2016hair}
W Guo and P Aarabi.
\newblock Hair segmentation using heuristically-trained neural networks.
\newblock {\em IEEE Transactions on Neural Networks and Learning Systems},
  29(1):25--36, 2016.

\bibitem{resnet}
Kaiming He, Xiangyu Zhang, Shaoqing Ren, and Jian Sun.
\newblock Deep residual learning for image recognition.
\newblock In {\em Proceedings of the IEEE conference on computer vision and
  pattern recognition}, pages 770--778, 2016.

\bibitem{celebahq}
Tero Karras, Timo Aila, Samuli Laine, and Jaakko Lehtinen.
\newblock Progressive growing of gans for improved quality, stability, and
  variation.
\newblock In {\em International Conference on Learning Representations}, 2018.

\bibitem{adaface}
Minchul Kim, Anil~K Jain, and Xiaoming Liu.
\newblock Adaface: Quality adaptive margin for face recognition.
\newblock In {\em Proceedings of the IEEE/CVF Conference on Computer Vision and
  Pattern Recognition}, pages 18750--18759, 2022.

\bibitem{klare2012face}
Brendan~F Klare, Mark~J Burge, Joshua~C Klontz, Richard W~Vorder Bruegge, and
  Anil~K Jain.
\newblock Face recognition performance: Role of demographic information.
\newblock {\em IEEE Transactions on information forensics and security},
  7(6):1789--1801, 2012.

\bibitem{labelme}
\url{https://github.com/wkentaro/labelme}.

\bibitem{le2015fast}
T.~Hoang~Ngan Le, Khoa Luu, and Marios Savvides.
\newblock Fast and robust self-training beard/moustache detection and
  segmentation.
\newblock In {\em 2015 International Conference on Biometrics (ICB)}, pages
  507--512, 2015.

\bibitem{le2012beard}
T~Hoang~Ngan Le, Khoa Luu, Keshav Seshadri, and Marios Savvides.
\newblock Beard and mustache segmentation using sparse classifiers on
  self-quotient images.
\newblock In {\em 2012 19th IEEE International Conference on Image Processing},
  pages 165--168. IEEE, 2012.

\bibitem{le2017semi}
T~Hoang~Ngan Le, Khoa Luu, Chenchen Zhu, and Marios Savvides.
\newblock Semi self-training beard/moustache detection and segmentation
  simultaneously.
\newblock {\em Image and Vision Computing}, 58:214--223, 2017.

\bibitem{levinshtein2018real}
Alex Levinshtein, Cheng Chang, Edmund Phung, Irina Kezele, Wenzhangzhi Guo, and
  Parham Aarabi.
\newblock Real-time deep hair matting on mobile devices.
\newblock In {\em 2018 15th Conference on Computer and Robot Vision (CRV)},
  pages 1--7. IEEE, 2018.

\bibitem{celeba}
Ziwei Liu, Ping Luo, Xiaogang Wang, and Xiaoou Tang.
\newblock Deep learning face attributes in the wild.
\newblock In {\em Proceedings of the IEEE International Conference on Computer
  Vision}, pages 3730--3738, 2015.

\bibitem{lu2019experimental}
Boyu Lu, Jun-Cheng Chen, Carlos~D Castillo, and Rama Chellappa.
\newblock An experimental evaluation of covariates effects on unconstrained
  face verification.
\newblock {\em IEEE Transactions on Biometrics, Behavior, and Identity
  Science}, 1(1):42--55, 2019.

\bibitem{MicrosoftFaceAPI}
\url{https://azure.microsoft.com/en-us/services/cognitive-services/face/}.

\bibitem{muhammad2018hair_figaro2}
Umar~Riaz Muhammad, Michele Svanera, Riccardo Leonardi, and Sergio Benini.
\newblock Hair detection, segmentation, and hairstyle classification in the
  wild.
\newblock {\em Image and Vision Computing}, 71:25--37, 2018.

\bibitem{nguyen2008image}
Minh~Hoai Nguyen, Jean-Francois Lalonde, Alexei~A. Efros, and Fernando De~la
  Torre.
\newblock Image-based shaving.
\newblock {\em Computer Graphics Forum}, 27(2):627--635, 2008.

\bibitem{deep_face}
Omkar~M. Parkhi, Andrea Vedaldi, and Andrew Zisserman.
\newblock Deep face recognition.
\newblock In {\em Proceedings of the British Machine Vision Conference (BMVC)},
  2015.

\bibitem{phillips1998feret}
P.Jonathon Phillips, Harry Wechsler, Jeffery Huang, and Patrick~J. Rauss.
\newblock The feret database and evaluation procedure for face-recognition
  algorithms.
\newblock {\em Image and Vision Computing}, 16(5):295--306, 1998.

\bibitem{morph}
Karl Ricanek~Jr and Tamirat Tesafaye.
\newblock Morph: A longitudinal image database of normal adult age-progression.
\newblock In {\em Proceedings of the 7th International Conference on Automatic
  Face and Gesture Recognition}, pages 341--345, 2006.

\bibitem{schroff2015facenet}
Florian Schroff, Dmitry Kalenichenko, and James Philbin.
\newblock Facenet: A unified embedding for face recognition and clustering.
\newblock In {\em Proceedings of the IEEE Conference on Computer Vision and
  Pattern Recognition}, pages 815--823, 2015.

\bibitem{serna2019algorithmic}
Ignacio Serna, Aythami Morales, Julian Fierrez, Manuel Cebrian, Nick
  Obradovich, and Iyad Rahwan.
\newblock Algorithmic discrimination: Formulation and exploration in deep
  learning-based face biometrics.
\newblock {\em arXiv preprint arXiv:1912.01842}, 2019.

\bibitem{seshadri2009robust}
Keshav Seshadri and Marios Savvides.
\newblock Robust modified active shape model for automatic facial landmark
  annotation of frontal faces.
\newblock In {\em 2009 IEEE 3rd International Conference on Biometrics: Theory,
  Applications, and Systems}, pages 1--8. IEEE, 2009.

\bibitem{shen2014image}
Yehu Shen, Zhenyun Peng, and Yaohui Zhang.
\newblock Image based hair segmentation algorithm for the application of
  automatic facial caricature synthesis.
\newblock {\em The Scientific World Journal}, 2014, 2014.

\bibitem{svanera2016figaro}
Michele Svanera, Umar~Riaz Muhammad, Riccardo Leonardi, and Sergio Benini.
\newblock Figaro, hair detection and segmentation in the wild.
\newblock In {\em 2016 IEEE International
  Cohttp://dictionary.cambridge.org/nference on Image Processing (ICIP)}, pages
  933--937. IEEE, 2016.

\bibitem{terhorst2021comprehensive}
Philipp Terh{\"o}rst, Jan~Niklas Kolf, Marco Huber, Florian Kirchbuchner, Naser
  Damer, Aythami~Morales Moreno, Julian Fierrez, and Arjan Kuijper.
\newblock A comprehensive study on face recognition biases beyond demographics.
\newblock {\em IEEE Transactions on Technology and Society}, 3(1):16--30, 2021.

\bibitem{thong2021feature}
William Thong and Cees~GM Snoek.
\newblock Feature and label embedding spaces matter in addressing image
  classifier bias.
\newblock In {\em Proceedings of the British Machine Vision Conference (BMVC)},
  2021.

\bibitem{wang2004face}
Haitao Wang, Stan~Z Li, and Yangsheng Wang.
\newblock Face recognition under varying lighting conditions using self
  quotient image.
\newblock In {\em Sixth IEEE International Conference on Automatic Face and
  Gesture Recognition, 2004. Proceedings.}, pages 819--824. IEEE, 2004.

\bibitem{cosface}
Hao Wang, Yitong Wang, Zheng Zhou, Xing Ji, Dihong Gong, Jingchao Zhou, Zhifeng
  Li, and Wei Liu.
\newblock Cosface: Large margin cosine loss for deep face recognition.
\newblock In {\em Proceedings of the IEEE Conference on Computer Vision and
  Pattern Recognitionn}, pages 5265--5274, 2018.

\bibitem{wu2022face}
Haiyu Wu, V{\'\i}tor Albiero, KS Krishnapriya, Michael~C King, and Kevin~W
  Bowyer.
\newblock Face recognition accuracy across demographics: Shining a light into
  the problem.
\newblock {\em arXiv preprint arXiv:2206.01881}, 2022.

\bibitem{haiyu}
Haiyu Wu, Grace Bezold, Aman Bhatta, and Kevin~W Bowyer.
\newblock Logical consistency and greater descriptive power for facial hair
  attribute learning.
\newblock {\em arXiv preprint arXiv:2302.11102}, 2023.

\bibitem{Xu_2021_CVPR}
Xingkun Xu, Yuge Huang, Pengcheng Shen, Shaoxin Li, Jilin Li, Feiyue Huang,
  Yong Li, and Zhen Cui.
\newblock Consistent instance false positive improves fairness in face
  recognition.
\newblock In {\em Proceedings of the IEEE/CVF Conference on Computer Vision and
  Pattern Recognition (CVPR)}, pages 578--586, June 2021.

\bibitem{yan2020two}
Yongzhe Yan, Stefan Duffner, Xavier Naturel, Anthony Berthelier, Christophe
  Garcia, Christophe Blanc, and Thierry Chateau.
\newblock Two-stage human hair segmentation in the wild using deep shape prior.
\newblock {\em Pattern Recognition Letters}, 136:293--300, 2020.

\bibitem{yoon2021real}
Ho-Sub Yoon, Seong-Woo Park, and Jang-Hee Yoo.
\newblock Real-time hair segmentation using mobile-unet.
\newblock {\em Electronics}, 10(2):99, 2021.

\bibitem{bisenet}
Changqian Yu, Jingbo Wang, Chao Peng, Changxin Gao, Gang Yu, and Nong Sang.
\newblock Bisenet: Bilateral segmentation network for real-time semantic
  segmentation.
\newblock In {\em Proceedings of the European conference on computer vision
  (ECCV)}, pages 325--341, 2018.

\end{thebibliography}
}

\end{document}